\documentclass[preprint,8pt]{elsarticle}
\usepackage{url}
\usepackage{pgfplots}
\usepackage{color,soul}
\soulregister\cite7
\soulregister\ref7
\soulregister\eqref7
\usepackage{mdframed}
\usepackage{amsmath}
\usepackage{amssymb}
\usepackage{amsthm}
\usepackage{enumerate}
\usepackage{blindtext}
\usepackage{multirow}
\usepackage{multicol}
\usepackage{graphicx}
\usepackage{epstopdf}
\usepackage{placeins}
\usepackage{caption}
\usepackage{algorithm}
\usepackage{algorithmicx}
\usepackage{pseudocode}
\usepackage{algpseudocode}
\usepackage{subcaption}
\usepackage{mathrsfs}

\DeclareMathOperator\vg{\vect{\gamma}}

\newcommand\dfdx[2]{\frac{\partial #1}{\partial #2}}
\newcommand\diag[1]{\mbox{diag}\left(#1\right)}

\DeclareMathOperator*{\argmin}{arg\,min}

\newcommand{\vect}[1]{\boldsymbol{#1}}
\newcommand{\iji}[1]{\left(#1\right)_{(i,j)}}

\newcommand{\vHijs}[1]{{H_{(i,j)}^{#1}}}
\newcommand{\vWijs}[1]{{W_{(i,j)}^{#1}}}

\newcommand{\vHs}[1]{{H^{#1}}}

\newcommand{\myexp}[2]{\left(#1\right)^{#2}}
\newcommand{\vHjs}[1]{{H_{(:,j)}^{#1}}}
\newcommand{\p}[1]{p\left(#1\right)}
\newcommand{\ps}[2]{p^{#1}\left(#2\right)}
\newcommand{\pr}[1]{p^R\left(#1\right)}
\newcommand{\Q}[1]{Q\left(#1\right)}
\newcommand{\vhs}[1]{\vect{h^s}}

\newcommand{\vHijts}[1]{{\bar{H}_{(i,j)}^{#1}}}
\newcommand{\Xj}[0]{X_{(:,j)}}
\newcommand{\vXj}[0]{{X_{(:,j)}}}
\newcommand{\vHj}[0]{\vect{H_{(:,j)}}}
\newcommand{\Hj}[0]{{H_{(:,j)}}}
\newcommand{\vH}[0]{\vect{H}}
\newcommand{\Hij}[0]{H_{(i,j)}}
\newcommand{\Wij}[0]{W_{(i,j)}}
\newcommand{\vHij}[0]{\vect{H_{(i,j)}}}
\newcommand{\vWij}[0]{{\vect{W}_{(i,j)}}}
\newcommand{\vHt}[0]{\bar{H}^t}
\newcommand{\Hts}[1]{{\bar{H}^{#1}}}

\newcommand{\vHijt}[0]{\bar{H}_{(i,j)}^t}

\newcommand{\vHjt}[1]{\bar{H}_{(:,j)}^{#1}}
\newcommand{\vgij}[0]{\vect{\gamma_{(i,j)}}}
\newcommand{\gij}[0]{\gamma_{(i,j)}}
\newcommand{\vgj}[0]{\vect{\gamma_{(:,j)}}}
\newcommand{\vOmt}[0]{{\Omega^t}}
\newcommand{\vOmijt}[0]{{\Omega_{(i,j)}^t}}
\newcommand{\vHgb}[0]{{H_{(g_b,j)}}}
\newcommand{\vHtgbj}[0]{{\bar{H}_{(g_b,j)}}}
\newcommand{\vHtij}[0]{\bar{H}_{(i,j)}}
\newcommand{\vSt}[0]{{S^t}}
\newcommand{\vOmjt}[0]{{\Omega_{(:,j)}^t}}

\newcommand{\RNum}[1]{\uppercase\expandafter{\romannumeral #1\relax}}

\newtheorem{theorem}{Theorem}

\newtheorem{corollary}{Corollary}

\DeclareMathOperator*{\R}{\mathbb{R}}
\newcommand\G[1]{\mathsf{\Gamma}\left( {#1} \right)}
\usepackage{xcolor}

\journal{Signal Processing}

\begin{document}

\begin{frontmatter}

\title{A Unified Framework for Sparse Non-Negative Least Squares using Multiplicative Updates and the Non-Negative Matrix Factorization Problem}



\author[mymainaddress]{Igor Fedorov\corref{mycorrespondingauthor}}

\cortext[mycorrespondingauthor]{Corresponding author}
\ead{ifedorov@eng.ucsd.edu}
\author[mymainaddress]{Alican Nalci}
\author[mysecondaryaddress]{Ritwik Giri}
\author[mymainaddress]{Bhaskar D. Rao}
\author[mymainaddress]{Truong Q. Nguyen}
\author[mymainaddress]{Harinath Garudadri}

\address[mymainaddress]{University of California, San Diego, 9500 Gilman Dr, San Diego, CA, 92103}
\address[mysecondaryaddress]{Starkey Hearing Technologies, 6700 Washington Avenue S.
Eden Prairie, MN 55344}

\begin{abstract}
We study the sparse non-negative least squares (S-NNLS) problem. S-NNLS occurs naturally in a wide variety of applications where an unknown, non-negative quantity must be recovered from linear measurements. We present a unified framework for S-NNLS based on a rectified power exponential scale mixture prior on the sparse codes. We show that the proposed framework encompasses a large class of S-NNLS algorithms and provide a computationally efficient inference procedure based on multiplicative update rules. Such update rules are convenient for solving large sets of S-NNLS problems simultaneously, which is required in contexts like sparse non-negative matrix factorization (S-NMF). We provide theoretical justification for the proposed approach by showing that the local minima of the objective function being optimized are sparse and the S-NNLS algorithms presented are guaranteed to converge to a set of stationary points of the objective function. We then extend our framework to S-NMF, showing that our framework leads to many well known S-NMF algorithms under specific choices of prior and providing a guarantee that a popular subclass of the proposed algorithms converges to a set of stationary points of the objective function. Finally, we study the performance of the proposed approaches on synthetic and real-world data.
\end{abstract}


\end{frontmatter}


\section{Introduction}
\label{section:introduction}
Least squares problems occur naturally in numerous research and application settings. At a high level, given an observation $x \in \mathbb{R}^d$ of $h \in \mathbb{R}^n$ through a linear system $W \in \mathbb{R}^{d \times n}$, the least squares problem refers to
\begin{align}\label{eq:least squares}
\argmin_h \Vert x - Wh \Vert_2^2.
\end{align}
Quite often, prior information about $h$ is known. For instance, $h$ may be known to be non-negative. Non-negative data occurs naturally in many applications, including text mining \cite{pauca2004text}, image processing \cite{monga2007robust}, speech enhancement \cite{loizou2005speech}, and spectral decomposition \cite{fevotte2009nonnegative}\cite{sajda2004nonnegative}. In this case, \eqref{eq:least squares} is modified to
\begin{align}\label{eq:nnls}
\argmin_{h \geq 0} \Vert x - Wh \Vert_2^2
\end{align}
where $h \geq 0$ refers to the elements of $h$ being constrained to be non-negative and \eqref{eq:nnls} is referred to as the non-negative least squares (NNLS) problem. A solution to \eqref{eq:nnls} can be obtained using the well-known active set Lawson-Hanson algorithm \cite{lawson1974solving} or one of its many variants \cite{bro1997fast}. In this work, we are interested in a specific flavor of NNLS problems where $n > d$. Under this constraint, the linear system in \eqref{eq:nnls} is underdetermined and admits an infinite number of solutions. To constrain the set of possible solutions, a sparsity constraint on $h$ can be added, leading to a sparse NNLS (S-NNLS) formulation:
\begin{align}\label{eq:snnls}
\argmin_{h \geq 0, \Vert h \Vert_0 \leq k} \Vert x - Wh \Vert_2^2
\end{align}
where $\Vert \cdot \Vert_0$ refers to the $\ell_0$ pseudo-norm, which counts the number of non-zero entries. Solving \eqref{eq:snnls} directly is difficult because the $\ell_0$ pseudo-norm is non-convex. In fact, solving \eqref{eq:snnls} requires a combinatorial search and has been shown to be NP-hard \cite{eladbook}. Therefore, greedy methods have been adopted to approximate the solution \cite{eladbook,eldar2012compressed}. One effective approach, called reverse sparse NNLS (rsNNLS) \cite{peharz2012sparse}, first finds an $h$ such that $\Vert x - W h \Vert_{2}^2 \leq \delta$ using the active-set Lawson-Hanson algorithm and then prunes $h$ with a greedy procedure until $\Vert h \Vert_0 \leq k$, all while maintaining $h \geq 0$. Other approaches include various relaxations of the $\ell_0$ pseudo-norm in \eqref{eq:snnls} using the $\ell_1$ norm \cite{hoyer2002non} or a combination of the $\ell_1$ and $\ell_2$ norms \cite{hoyer2004non}, leading to easier optimization problems.

The purpose of this work is to address the S-NNLS problem in a setting often encountered by practitioners, i.e. when several S-NNLS problems must be solved simultaneously. We are primarily motivated by the problem of sparse non-negative matrix factorization (S-NMF). NMF falls under the category of dictionary learning algorithms. Dictionary learning is a common ingredient in many signal processing and machine learning algorithms \cite{mairal2009supervised,tovsic2011dictionary,gangeh2015supervised,kreutz2003dictionary}. In NMF, the data, the dictionary, and the encoding of the data under the dictionary are all restricted to be non-negative. Constraining the encoding of the data to be non-negative leads to the intuitive interpretation of the data being decomposed into an additive combination of dictionary atoms \cite{lee2001algorithms,aharon2005k,lee1999learning}. More formally, let $X \in \R^{d \times m}_+$ be a matrix representing the given data, where each column of $X$, $\vXj \in \R^d_+, 1 \leq j \leq m$, is a data vector. The goal of NMF is to decompose $X$ into two matrices $W \in {\R}^{d \times n}_+$ and $H \in {\R}^{n \times m}_+$. When $n < d$, NMF is often stated in terms of the optimization problem
\begin{align}\label{eq:NMF}
\theta^* = \argmin_{\theta \geq 0} \Vert X - W H \Vert_F^2
\end{align}
where $\theta = \lbrace W,H\rbrace$, $W$ is called the dictionary, $H$ is the encoding of the data under the dictionary, and $\theta \geq 0$ is short-hand for the elements of $W$ and $H$ being constrained to be non-negative. Optimizing \eqref{eq:NMF} is difficult because it is not convex in $\theta$ \cite{lin2007projected}. Instead of performing joint optimization, a block coordinate descent method \cite{bertsekas1999nonlinear} is usually adopted where the algorithm alternates between holding $W$ fixed while optimizing $H$ and vice versa \cite{lee2001algorithms,lee1999learning,lin2007projected,zhou2012fast,zhou2014nonnegative}:
\begin{align}
\text{Update } W \text{ given } H \label{eq:update w given h} \\
\text{Update } H \text{ given } W \label{eq:update h given w}.
\end{align}
Note that \eqref{eq:update w given h} and \eqref{eq:update h given w} are a collection of $d$ and $m$ NNLS problems, respectively, which motivates the present work. The block coordinate descent method is advantageous because \eqref{eq:update w given h} and \eqref{eq:update h given w} are convex optimization problems for the objective function in \eqref{eq:NMF}, so that any number of  techniques can be employed within each block. One of the most widely used optimization techniques, called the multiplicative update rules (MUR's), performs \eqref{eq:update w given h}-\eqref{eq:update h given w} using simple element-wise operations on $W$ and $H$ \cite{lee2001algorithms,lee1999learning}:
\begin{align}
\label{eq:MUR W}
W^{t+1} &= W^{t} \odot \frac{X H^T}{W^t H H^T}\\
\label{eq:MUR H}
H^{t+1} &= H^t \odot \frac{W^T X}{W^T W H^t}
\end{align}
where $\odot$ denotes element-wise multiplication, $A/B$ denotes element-wise division of matrices $A$ and $ B$, and $t$ denotes the iteration index. The MUR's shown in \eqref{eq:MUR W}-\eqref{eq:MUR H} are guaranteed to not increase the objective function in \eqref{eq:NMF} \cite{lee2001algorithms,lee1999learning} and, due to their simplicity, are widely used in the NMF community \cite{kim2015mixtures,joder2012real,raj2010non}. The popularity of NMF MUR's persists despite the fact that there is no guarantee that the sequence $\lbrace W^t,H^t\rbrace_{t=0}^\infty$ generated by \eqref{eq:MUR W}-\eqref{eq:MUR H} will converge to a local minimum \cite{gonzalez2005accelerating} or even a stationary point \cite{lin2007projected,gonzalez2005accelerating} of \eqref{eq:NMF}.

Unlike traditional NMF methods \cite{lee2001algorithms,lee1999learning}, this work considers the scenario where $W$ is overcomplete, i.e. $n \gg d$. Overcomplete dictionaries have much more flexibility to represent diverse signals \cite{donoho2006stable} and, importantly, lead to effective sparse and low dimensional representations of the data \cite{aharon2005k,donoho2006stable}. As in NNLS, the concept of sparsity has an important role in NMF because when $W$ is overcomplete, \eqref{eq:NMF} is not well-posed without some additional regularization. Sparsity constraints limit the set of possible solutions of \eqref{eq:NMF} and, in some cases, lead to guarantees of uniqueness \cite{gillis2012sparse}. The S-NMF problem can be stated as the solution to
\begin{align}\label{eq:l0 NMF}
\theta^* &= \argmin_{\theta \geq 0, \Vert H \Vert_0 \leq k} \Vert X - W H \Vert_F^2
\end{align}
where $\Vert H \Vert_0 \leq k$ is shorthand for $\lbrace \Vert \Hj \Vert_0 \leq k\rbrace_{j=1}^m $. One classical approach to S-NMF relaxes the $\ell_0$ constraint and appends a convex, sparsity promoting $\ell_1$ penalty to the objective function \cite{hoyer2002non}:
\begin{align}\label{eq:l1 NMF}
\theta^* &= \argmin_{\theta \geq 0} \Vert X - W H \Vert_F^2 + \lambda \Vert H \Vert_1
\end{align}
where $\Vert H \Vert_1$ is shorthand for $\sum_{j=1}^m \Vert \Hj \Vert_1$. As shown in \cite{hoyer2002non}, \eqref{eq:l1 NMF} can be iteratively minimized through a sequence of multiplicative updates where the update of $W$ is given by \eqref{eq:MUR W} and the update of $H$ is given by
\begin{align}\label{eq:MUR H l1}
H^{t+1} &= H^t \odot \frac{W^T X}{W^T W H^t+\lambda}.
\end{align}
We also consider an extension of S-NMF where a sparsity constraint is placed on $W$ \cite{hoyer2004non}
\begin{align}\label{eq:l0 NMF-W}
\theta^* &= \argmin_{\theta \geq 0, \Vert H \Vert_0 \leq k_h,\Vert W \Vert_0 \leq k_w} \Vert X - W H \Vert_F^2
\end{align}
which encourages basis vectors that explain localized features of the data \cite{hoyer2004non}. We refer to \eqref{eq:l0 NMF-W} as S-NMF-W.

The motivation of this work is to develop a maximum a-posteriori (MAP) estimation framework to address the S-NNLS and S-NMF problems. We build upon the seminal work in \cite{tipping2001sparse} on Sparse Bayesian Learning (SBL). The SBL framework places a sparsity-promoting prior on the data \cite{wipf2004sparse} and has been shown to give rise to many models used in the compressed sensing literature \cite{wipf2010iterative}. It will be shown that the proposed framework provides a general class of algorithms that can be tailored to the specific needs of the user. Moreover, inference can be done through a simple MUR for the general model considered and the resulting S-NNLS algorithms admit convergence guarantees.

The key contribution of this work is to detail a unifying framework that encompasses a large number of existing S-NNLS and S-NMF approaches. Therefore, due to the very nature of the framework, many of the algorithms presented in this work are not new. Nevertheless, there is value in the knowledge that many of the algorithms employed by researchers in the S-NNLS and S-NMF fields are actually members of the proposed family of algorithms. In addition, the proposed framework makes the process of formulating novel task-specific algorithms easy. Finally, the theoretical analysis of the proposed framework applies to any member of the family of proposed algorithms. Such an analysis has value to both existing S-NNLS and S-NMF approaches like \cite{lefevre2011itakura,grady2008compressive}, which do not perform such an analysis, as well as to any future approaches which fall under the umbrella of the proposed framework. It should be noted that several authors have proposed novel sets of MUR's with provable convergence guarantees for the NMF problem in \eqref{eq:NMF}  \cite{lin2007convergence} and S-NMF problem in \eqref{eq:l1 NMF} \cite{zhao2016unified}. In contrast to \cite{zhao2016unified}, the proposed framework does not use the $\ell_1$ regularization function to solve \eqref{eq:l0 NMF}. In addition, since the proposed framework encompasses the update rules used in existing works, the analysis presented here applies to works from existing literature, including \cite{lefevre2011itakura,grady2008compressive}.

\subsection{Contributions of the Paper}
\label{section:contributions}
\begin{itemize}
\item A general class of rectified sparsity promoting priors is presented and it is shown that the computational burden of the resulting inference procedure is handled by a class of simple, low-complexity MUR's.
\item A monotonicity guarantee for the proposed class of MUR's is provided, justifying their use in S-NNLS and S-NMF algorithms.
\item A convergence guarantee for the proposed class of S-NNLS and S-NMF-W algorithms is provided.
\end{itemize}

\subsection{Notation}
Bold symbols are used to denote random variables and plain font to denote a particular realization of a random variable. MATLAB notation is used to denote the $(i,j)$'th element of the matrix $H$ as $\Hij$ and the $j$'th column of $H$ as $\Hj$. We use $\vHs{s}$ to denote the matrix $H$ at iteration $s$ of a given algorithm and $\myexp{H}{z}$ to denote the matrix $H$ with each element raised to the power $z$.

\section{Sparse Non-Negative Least Squares Framework Specification}

\begin{table*}
\centering
\begin{tabular}{| c  c |}
\hline
Distribution & pdf \\  \hline
Rectified Gaussian & $\ps{\mathsf{RG}}{h;\gamma} =  \sqrt{\dfrac{2}{\pi \gamma}}\exp \left(-\dfrac{h^2}{2\gamma}\right) u(h)$ \\ 
Exponential & $\ps{\mathsf{Exp}}{h;\gamma} = \gamma \exp\left(-\gamma h\right) u(h)$ \\ 
Inverse Gamma & $\ps{\mathsf{IGa}}{h;a,b} = \frac{b^a}{\G{a}} h^{-a-1} \exp \left(-\frac{b}{h}\right) u(h)$ \\
Gamma & $\ps{\mathsf{Ga}}{h;a,b} = \frac{1}{\G{a}b^a}h^{a-1}\exp\left(-hb\right) u(h)$\\ 
Rectified Student's-t  & $\ps{\mathsf{RST}}{h; \tau} = \frac{2\mathsf{\Gamma}\left(\frac{\tau+1}{2} \right)}{\sqrt{\tau \pi} \G{\frac{\tau}{2}}}\left( 1+ \frac{h^2}{\tau}\right)^{-\frac{\left(\tau+1\right)}{2}} u(h)$\\
Rectified Generalized Double Pareto  & $\ps{\mathsf{RGDP}}{h; a,b,\tau} = 2 \eta \left(1+\frac{ h^b}{\tau a^b} \right)^{-\left(\tau+\frac{1}{b}\right)} u(h)$ \\\hline
\end{tabular}
\caption{Distributions used throughout this work, where $\exp\left(a\right) = e^{a}$.}
\label{table:distributions}
\end{table*}

\label{section:Bayesian sparse NMF}
The S-NNLS signal model is given by
\begin{align}\label{eq:signal model}
\vect{X} = \vect{W H} + \vect{V}
\end{align}
where the columns of $\vect{V}$, the noise matrix, follow a $\mathsf{N}(0,\sigma^2 \mathsf{I})$ distribution. To complete the model, a prior on the columns of $\vH$, which are assumed to be independent and identically distributed, must be specified. This work considers separable priors of the form $\p{\Hj} = \prod_{i=1}^n \p{\Hij}$, where $\p{\Hij}$ has a scale mixture representation \cite{andrews1974scale,RGSM}:
\begin{align}\label{eq:scale mixture}
\p{\Hij} = \int_{0}^\infty \p{\Hij | \gij} \p{\gij} d\gij.
\end{align}
Separable priors are considered because, in the absence of
prior knowledge, it is reasonable to assume independence
amongst the coefficients of $\vH$. The case where dependencies amongst the coefficients exist is considered in Section \ref{section:block sparse}. The proposed framework extends the work on power exponential scale mixtures \cite{giri2015type,ritwikMagazine} to rectified priors and uses the Rectified Power Exponential (RPE) distribution for the conditional density of $\vHij$ given $\vgij$:
\begin{align*}
\ps{\mathsf{RPE}}{\Hij | \gij;z} = \frac{z{e^{-\left(\frac{\Hij}{\gij} \right)}}^z}{\gij \G{\frac{1}{z}}} u(\Hij)
\end{align*}
where $u(\cdot)$ is the unit-step function, $0 < z \leq 2$, and $\mathsf{\Gamma}(a) = \int_{0}^\infty t^{a-1} e^{-t} dt$.
The RPE distribution is chosen for its flexibility. In this context, \eqref{eq:scale mixture} is referred to as a rectified power exponential scale mixture (RPESM).

The advantage of the scale mixture prior is that it introduces a Markovian structure of the form
\begin{align}\label{eq:Makovian}
\vgj \rightarrow \vHj \rightarrow \vect{\vXj}
\end{align}
and inference can be done in either the $\vH$ or $\vg$ domains. This work focuses on doing MAP inference in the $\vH$ domain, which is also known as Type 1 inference, whereas inference in the $\vg$ domain is referred to as Type 2. The scale mixture representation is flexible enough to represent most heavy-tailed densities \cite{palmer2006variational,palmer2005variational,lange1993normal,dempster1980iteratively,dempster1977maximum}, which are known to be the best sparsity promoting priors \cite{tipping2001sparse,wipf2007empirical}. One reason for the use of heavy-tailed priors is that they are able to model both the sparsity and large non-zero entries of $\vH$.

The RPE encompasses many rectified distributions of interest. For instance, the RPE reduces to a Rectified Gaussian by setting $z = 2$, which is a popular prior for modeling non-negative data \cite{schachtner2014bayesian,RGSM} and results in a Rectified Gaussian Scale Mixture in \eqref{eq:scale mixture}. Setting $z = 1$ corresponds to an Exponential distribution and leads to an Exponential Scale Mixture in \eqref{eq:scale mixture} \cite{themelis2012novel}. Table \ref{table:rpesm} shows that many rectified sparse priors of interest can be represented as a RPESM. Distributions of interest are summarized in Table \ref{table:distributions}.
\begin{table}
\centering
\begin{tabular}{ccc}
\hline
$z$ & $\ps{}{\gij}$ & $ \ps{}{\Hij}$ \\ \hline 
$2$ & $\ps{\mathsf{Exp}}{\gij ; \tau^2/2}$ & $\ps{\mathsf{Exp}}{\Hij ; \tau}$ \\
$2$ & $\ps{\mathsf{Exp}}{\gij ; \tau^2/2}$ & $\ps{\mathsf{RST}}{\Hij ; \tau}$ \\ 
$1$ & $\ps{\mathsf{Ga}}{\gij;\tau,\tau}$ & $\ps{\mathsf{{RGDP}}}{\Hij; 1, 1 , \tau}$ \\ \hline
\end{tabular}
\caption{RPESM representation of rectified sparse priors.}
\label{table:rpesm}
\end{table}

\section{Unified MAP Inference Procedure}
\label{section:Type 1}
In the MAP framework, $H$ is directly estimated from $X$ by minimizing
\begin{align}
L(H) &= -\log \left( \prod_{j=1}^m \p{\Hj | \Xj}\right)\label{eq:Type1}.
\end{align}
We have made the dependence of the negative log-likelihood on $X$ and $W$ implicit for brevity. Minimizing \eqref{eq:Type1} in closed form is intractable for most priors, so the proposed framework resorts to an Expectation-Maximization (EM) approach \cite{dempster1977maximum}. In the E-step, the expectation of the negative complete data log-likelihood with respect to the distribution of $\vg$, conditioned on the remaining variables, is formed:
\begin{align}\label{eq:Type 1 Q function}
\begin{split}
&\Q{H,\vHt} \dot{=} \Vert X - W H \Vert_F^2 + \lambda \left(\sum_{i=1,j=1}^{i=n,j=m} \myexp{\Hij}{z} \left\langle \dfrac{1}{\left(\gamma_{(i,j)}\right)^{z}} \right\rangle - \log u\left(\Hij\right)\right)
\end{split}
\end{align}
where $\langle \cdot \rangle$ refers to the expectation with respect to the density $\p{\gamma_{(i,j)} | \vHijt}$, $t$ refers to the iteration index, $\vHt$ denotes the estimate of $H$ at the $t$'th EM iteration, and $\dot{=}$ refers to dropping terms that do not influence the M-step and scaling by $\lambda = 2\sigma^2$. The last term in \eqref{eq:Type 1 Q function} acts as a barrier function against negative values of $H$. The function $\Q{H,\vHt}$ is separable in the columns of $H$. In an abuse of notation, we use $\Q{\Hj,\vHjt{t}}$ to refer to the dependency of $\Q{H,\vHt}$ on $\Hj$.

In order to compute the expectation in \eqref{eq:Type 1 Q function}, a similar method to the one used in \cite{giri2015type,palmer2006variational} is employed, with some minor adjustments due to non-negativity constraints. Let $\p{\Hij} = \pr{\Hij}u\left(\Hij\right)$, where $\pr{\Hij}$ is the portion of $\p{\Hij}$ that does not include the rectification term, and let $\pr{\Hij}$ be differentiable on $[0,\infty)$. Then,
\begin{align}\label{eq:gamma expectation}
 \left\langle \dfrac{1}{\left(\gamma_{(i,j)}\right)^{z}} \right\rangle = -\dfdx{\log \pr{\vHijt}}{\vHijt} \frac{1}{z\left(\vHijt\right)^{z-1}}.
\end{align}
Turning to the M-step, the proposed approach employs the Generalized EM (GEM) M-step \cite{dempster1977maximum}:
\begin{align}
&\text{Choose } \Hts{t+1} \text{ such that } Q(\Hts{t+1},\vHt) \leq Q(\vHt,\vHt) \label{eq:Type 1 GEM M-step} \tag{GEM M-step}.
\end{align}
In particular, $Q(H,\vHt)$ is minimized through an iterative gradient descent procedure. As with any gradient descent approach, selection of the learning rate is critical in order to ensure that the objective function is decreased and the problem constraints are met. Following \cite{lee1999learning,lee2001algorithms}, the learning rate is selected such that the gradient descent update is guaranteed to generate non-negative updates and can be implemented as a low-complexity MUR, given by
\begin{align}\label{eq:mult rule nmf}
\vHs{s+1} &= \vHs{s} \odot \frac{W^T X}{W^T W H^s + \lambda \vOmt \odot \myexp{\vHs{s}}{z-1}}\\
\vOmijt &=  -\frac{1}{\myexp{\vHijt}{z-1}} \dfdx{\log \pr{\vHijt}}{\vHijt} \nonumber
\end{align}
where $s$ denotes the gradient descent iteration index (not to be confused with the EM iteration index $t$). The resulting S-NNLS algorithm is summarized in Algorithm \ref{alg:Type 1}, where $\zeta$ denotes the specific MUR used to update $H$, which is \eqref{eq:mult rule nmf} in this case.
\begin{algorithm}
\caption{S-NNLS Algorithm}
\label{alg:Type 1}
\begin{algorithmic}
\Require $X,W,\Hts{0},\lambda,\zeta,S,t^\infty$
\State Initialize $t = 0,\mathscr{Z} = \lbrace (i,j) \rbrace_{i=1,j=1}^{i=n,j=m}$
\While{$\mathscr{Z} \neq \emptyset$}
\State Form $\vOmt$ and initialize $\vHs{1} = \vHt,\mathscr{J} = \mathscr{Z}$
\For{$s=1$ to $S$}
\State Generate $\vHijs{s+1}$ using update rule $\zeta$ for $(i,j) \in \mathscr{J}$
\State Set $\vHijs{s+1} = \vHijs{s}$ for any $(i,j) \notin \mathscr{J}$
\State $\mathscr{J} \leftarrow \mathscr{J} \setminus \left\lbrace (i,j): \vHijs{s+1} = 0 \text{ or } \vHijs{s+1} = \vHijs{s}\right\rbrace$
\EndFor
\State Set ${\bar{H}^{t+1}} = \vHs{S+1}$ and $\mathscr{Z} \leftarrow \mathscr{Z} \setminus \left\lbrace (i,j) : \vHijts{t+1} = \vHijts{t} \text{ or } \vHijts{t+1} = 0\right\rbrace$
\State $t \leftarrow t+1$
\If{$t = t^\infty$}
\State Break
\EndIf
\EndWhile
\State Return $\bar{H}^{t}$
\end{algorithmic}
\end{algorithm}

\subsection{Extension to S-NMF}
We now turn to the extension of our framework to the S-NMF problem. As before, the signal model in \eqref{eq:signal model} is used as well as the RPESM prior on $H$. To estimate $W$ and $H$, the proposed framework seeks to find
\begin{align}\label{eq:nmf likelihood}
\argmin_{W,H} L^{NMF}(W,H), \; \; L^{NMF}(W,H) = -\log p(W,H | X).
\end{align}
The random variables $\vect{W}$ and $\vect{H}$ are assumed independent and a non-informative prior over the positive orthant is placed on $\vect{W}$ for S-NMF. For S-NMF-W, a separable prior from the RPESM family is assumed for $\vect{W}$. In order to solve \eqref{eq:nmf likelihood},   the block-coordinate descent optimization approach in \eqref{eq:update w given h}-\eqref{eq:update h given w} is employed. For each one of \eqref{eq:update w given h} and \eqref{eq:update h given w}, the GEM procedure described above is used.

The complete S-NMF/S-NMF-W algorithm is given in Algorithm  \ref{alg:Sparse NMF}. Due to the symmetry between \eqref{eq:update w given h} and \eqref{eq:update h given w} and to avoid unnecessary repetition, heavy use of Algorithm \ref{alg:Type 1} in Algorithm \ref{alg:Sparse NMF} is made. Note that $\zeta_h = $ \eqref{eq:mult rule nmf}, $\zeta_w = $ \eqref{eq:MUR H} for S-NMF, and $\zeta_w = $ \eqref{eq:mult rule nmf} for S-NMF-W.

\begin{algorithm}
\caption{S-NMF/S-NMF-W Algorithm}
\label{alg:Sparse NMF}
\begin{algorithmic}
\Require $X,\lambda,S,\zeta_w,\zeta_h,t^\infty$
\State Initialize $\vWijs{0} = 1, \vHijs{0}=1, t = 0$
\While{$t \neq t^\infty$ and $\left(\bar{H}^{t+1} \neq \bar{H}^{t} \text{ or } \bar{W}^{t+1} \neq \bar{W}^{t}\right)$}
\State \small $\bar{W}^{t+1} = \left(\text{Algorithm1}(X^T,\left(\bar{H}^{t}\right)^T,\left(\bar{W}^{t}\right)^T,\lambda,\zeta_w,S,1)\right)^T$
\State $\Hts{t+1} = \text{Algorithm1}(X,\bar{W}^{t+1},\bar{H}^{t},\lambda,\zeta_h,S,1)$
\State $t \leftarrow t+1$
\EndWhile
\end{algorithmic}
\end{algorithm}

\section{Examples of S-NNLS and S-NMF Algorithms}
\label{section:examples}
In the following, evidence of the utility of the proposed framework is provided by detailing several specific algorithms which naturally arise from \eqref{eq:mult rule nmf} with different choices of prior. It will be shown that the algorithms described in this section are equivalent to well-known S-NNLS and S-NMF algorithms, but derived in a completely novel way using the RPESM prior. The S-NMF-W algorithms described are, to the best of our knowledge, novel. In Section \ref{section:block sparse}, it will be shown that the proposed framework can be easily used to define novel algorithms where block-sparsity is enforced.

\subsection{Reweighted $l_2$}
\label{section:S-NNLS rl2}
Consider the prior $\vHij \sim \ps{\mathsf{RST}}{\Hij ; \tau}$. Given this prior, \eqref{eq:mult rule nmf} becomes
\begin{align}\label{eq:mult rule rl2}
\vHs{s+1} = \vHs{s} \odot \dfrac{W^T X}{W^T W \vHs{s} +  \frac{2\lambda\left(\tau+1\right)\vHs{s}}{\tau + \left(\vHt\right)^2}}.
\end{align}
Given this choice of prior on $\vHij$ and a non-informative prior on $\vWij$, it can be shown that $L^{NMF}(W,H)$ reduces to
\begin{align}\label{eq:rl2 obj func}
\Vert X-W H\Vert_F^2 +\tilde{\lambda}\sum_{i=1,j=1}^{i=n,j=m} \log\left(\myexp{\Hij}{2}+\tau\right)
\end{align}
over $H \in \mathbb{R}_+^{n\times m}$ and $W \in \mathbb{R}_{+}^{d\times n}$ (i.e. the $\log u(\cdot)$ terms have been omitted for brevity), where $\tilde{\lambda} = 2\sigma^2\left(\tau+1\right)$. The sparsity-promoting regularization term in \eqref{eq:rl2 obj func} was first studied in \cite{chartrand2008iteratively} in the context of vector sparse coding (i.e. without non-negativity constraints). Majorizing the sparsity promoting term in \eqref{eq:rl2 obj func}, it can be shown that \eqref{eq:rl2 obj func} is upper-bounded by 
\begin{align}\label{eq:rl2 nuirls}
\Vert X-W H\Vert_2^2 +\tilde{\lambda} \left\Vert \frac{H}{Q^t} \right\Vert_F^2
\end{align}
where $Q_{(i,j)}^t = \bar{H}^t_{(i,j)}+\tau$. Note that this objective function was also used in \cite{chen2013sparse}, although it was optimized using a heuristic approach based on the Moore-Penrose pseudoinverse operator. Letting $R = H/Q^t$ and $\tilde{\lambda} \rightarrow 0$, \eqref{eq:rl2 nuirls} becomes
\begin{align}\label{eq:nuirls obj}
\Vert X - W \left( Q^t \odot R\right) \Vert_2^2 
\end{align}
which is exactly the objective function that is iteratively minimized in the NUIRLS algorithm \cite{grady2008compressive} if we let $\tau \rightarrow 0$. Although \cite{grady2008compressive} gives a MUR for minimizing \eqref{eq:nuirls obj}, the MUR can only be applied for each column of $H$ individually. It is not clear why the authors of \cite{grady2008compressive} did not give a matrix based update rule for minimizing \eqref{eq:nuirls obj}, which can be written as 
\begin{align*}
R^{s+1} = R^s \odot \dfrac{W^T X}{W^T W \left(Q^t \odot R^{s}\right)}.
\end{align*}
This MUR is identical to \eqref{eq:mult rule rl2} in the setting $\lambda,\tau \rightarrow 0$. Although \cite{grady2008compressive} makes the claim that NUIRLS converges to a local minimum of \eqref{eq:nuirls obj}, this claim is not proved. Moreover, nothing is said regarding convergence with respect to the actual objective function being minimized (i.e. \eqref{eq:rl2 obj func} as opposed to the majorizing function in \eqref{eq:nuirls obj}). As the analysis in Section \ref{section:analysis} will reveal, using the update rule in \eqref{eq:mult rule rl2} within Algorithm \ref{alg:Type 1}, the iterates are guaranteed to converge to a stationary point of \eqref{eq:rl2 obj func}. We make no claims regarding convergence with respect to the majorizing function in \eqref{eq:rl2 nuirls} or \eqref{eq:nuirls obj}.

\subsection{Reweighted $\ell_1$}
\label{section: S-NNLS rl1}
Assuming $\vHij \sim \ps{\mathsf{RGDP}}{\Hij;1,1,\tau}$, \eqref{eq:mult rule nmf} reduces to
\begin{align}\label{eq:mult rule rl1}
\vHs{s+1} = \vHs{s} \odot \dfrac{W^T X}{W^T W \vHs{s} +  \frac{\lambda(\tau+1)}{\tau+\vHt}}.
\end{align}
Plugging the RGDP prior into \eqref{eq:nmf likelihood} and assuming a non-informative prior on $\vWij$ leads to the Lagrangian of the objective function considered in \cite{candes2008enhancing} for unconstrained vector sparse coding (after omitting the barrier function terms): $\Vert X - W H \Vert_F^2 + \tilde{\lambda}\sum_{i=1,j=1}^{i=n,j=m}\log(\Hij+\tau)$. Interestingly, this objective function is a special case of the block sparse objective considered in \cite{lefevre2011itakura} (where the Itakura-Saito reconstruction loss is used instead of the Frobenius norm loss) if each $\Hij$ is considered a separate block. The authors of \cite{lefevre2011itakura} did not offer a convergence analysis of their algorithm, in contrast with the present work. To the best of our knowledge, the reweighted $\ell_1$ formulation has not been considered in the S-NNLS literature.

\subsection{Reweighted $\ell_2$ and Reweighted $\ell_1$ for S-NMF-W}
Using the reweighted $\ell_2$ or reweighted $\ell_1$ formulations to promote sparsity in $W$ is straightforward in the proposed framework and  involves setting $\zeta_w$ to \eqref{eq:mult rule rl2} or \eqref{eq:mult rule rl1}, respectively, in Algorithm \ref{alg:Sparse NMF}.

\section{Extension to Block Sparsity}
\label{section:block sparse}
As a natural extension of the proposed framework, we now consider the block sparse S-NNLS problem. This section will focus on the S-NNLS context only because the extension to S-NMF is straightforward. Block sparsity arises naturally in many contexts, including speech processing \cite{kim2015mixtures,sun2013universal}, image denoising \cite{dong2011sparsity}, and system identification \cite{jiang2014block}. The central idea behind block-sparsity is that $W$ is assumed to be divided into disjoint blocks and each $\vXj$ is assumed to be a linear combination of the elements of a \textit{small number of blocks}. This constraint can be easily accommodated by changing the prior on $\vHj$ to a block rectified power exponential scale mixture:
\begin{align}\label{eq:block sparse prior}
\begin{split}
\p{\Hj} &= \prod_{g_b \in \mathscr{G}} \underbrace{ \int_{0}^\infty \prod_{i \in g_b} \p{\Hij | \gamma_{(b,j)}} \p{\gamma_{(b,j)}} d\gamma_{(b,j)}}_{\p{\vHgb}}\\
\end{split}
\end{align}
where $\mathscr{G}$ is a disjoint union of $\lbrace g_b \rbrace_{b=1}^B$ and $\vHgb$ is a vector consisting of the elements of $\Hj$ whose indices are in $g_b$. To find the MAP estimate of $H$ given $X$, the same GEM procedure as before is employed, with the exception that the computation of the weights in \eqref{eq:Type 1 Q function} is modified to:
\begin{align*}
\left\langle \frac{1}{\left(\gamma_{(b,j)}\right)^z}\right\rangle = -\dfdx{\log \pr{\vHtgbj}}{\vHtij} \frac{1}{z\myexp{\vHtij}{z-1}}
\end{align*}
where $i \in g_b$. It can be shown that the MUR for minimizing $Q(H,\vHt)$ in \eqref{eq:mult rule nmf} can be modified to account for the block prior in \eqref{eq:block sparse prior} to
\begin{align}\label{eq:mult rule block sparse nmf}
H^{s+1} &= H^s \odot \frac{W^T X}{W^T W H^s + \lambda {\Phi^t} \odot \myexp{\vHs{s}}{z-1}}\\
{\Phi_{(g_b,j)}^t} &=  -\frac{1}{\myexp{\vHijt}{z-1}} \dfdx{\log \pr{{\bar{H}_{(g_b,j)}^t}}}{\vHijt} \text{ for any $i \in g_b$}.\nonumber
\end{align}
Next, we show examples of block S-NNLS algorithms that arise from our framework.

\subsection{Example: Reweighted $\ell_2$ Block S-NNLS}
Consider the block-sparse prior in \eqref{eq:block sparse prior}, where $\p{\Hij | \gamma_{(b,j)}}$, $i \in g_b$, is a RPE with $z=2$ and $\vect{\gamma_{(b,j)}} \sim \ps{\mathsf{IGa}}{\gamma_{(b,j)};{\tau}/{2},{\tau}/{2}}$. The resulting density $\p{\Hj}$ is a block RST (BRST) distribution:
\begin{align*}
\p{\Hj} = \left(\prod_{g_b \in \mathscr{G}} \frac{2\G{\frac{\tau+1}{2}}}{\sqrt{\pi \tau}\G{\frac{\tau}{2}}}\left(1+\frac{\Vert \vHgb \Vert_2^2}{\tau}\right)^{-\frac{\left(\tau+1\right)}{2}}\right)\prod_{i=1}^n u\left(\Hij\right).
\end{align*}
The MUR for minimizing $Q(H,\vHt)$ under the BRST prior is given by:
\begin{align}\label{eq:mult rule block rl2}
\vHs{s+1} &= \vHs{s} \odot \frac{W^T X}{W^T W\vHs{s} + \frac{2\lambda (\tau+1) \vHs{s}}{\tau + \vSt}}
\end{align}
where ${S_{(g_b,j)}^t} = \Vert {\bar{H}_{(g_b,j)}^t} \Vert_2^2$.

\subsection{Example: Reweighted $\ell_1$ Block S-NNLS}
Consider the block-sparse prior in \eqref{eq:block sparse prior}, where $\p{\Hij |\gamma_{(b,j)}}$, $i \in g_b$, is a RPE with $z = 1$ and $\vect{\gamma_{(b,j)}} \sim \ps{\mathsf{Ga}}{\gamma_{(b,j)};\tau,\tau}$. The resulting density $\p{\Hj}$ is a block rectified generalized double pareto (BRGDP) distribution:
\begin{align*}
\p{\Hj} = \left(\prod_{g_b \in \mathscr{G}} {2\eta}\myexp{1+\frac{\Vert \vHgb \Vert_1}{\tau}}{-\left(\tau+1\right)}\right)\prod_{i=1}^n u\left(\Hij\right).
\end{align*}
The MUR for minimizing $Q(H,\vHt)$ under the BRGDP prior is given by:
\begin{align}\label{eq:mult rule block rl1}
H^{t+1} &= \vHt \odot \frac{W^T X}{W^T W\vHt + \frac{\lambda(\tau+1)}{\tau+V^t}}
\end{align}
where ${V_{(g_b,j)}^t} = \Vert {\bar{H}_{(g_b,j)}^t} \Vert_1$.

\subsection{Relation To Existing Block Sparse Approaches}
Block sparse coding algorithms are generally characterized by their block-sparsity measure. The analog of the $\ell_0$ sparsity measure for block-sparsity is the $\ell_2 - \ell_0$ measure $
\sum_{g_b \in \mathscr{G}} 1_{\Vert \vHgb \Vert_2 > 0}$, which simply counts the number of blocks with non-zero energy. This sparsity measure has been studied in the past and block versions of the popular MP and OMP algorithms have been extended to Block-MP (BMP) and Block-OMP (BOMP) \cite{eldar2010block}. Extending BOMP to non-negative BOMP (NNBOMP) is straightforward, but details are omitted due to space considerations. One commonly used block sparsity measure in the NMF literature is the $\log-\ell_1$ measure \cite{lefevre2011itakura}: $
\sum_{g_b \in \mathscr{G}} \log(\Vert \vHgb \Vert_1+\tau)$.
This sparsity measure arises naturally in the proposed S-NNLS framework when the BRGDP prior is plugged into \eqref{eq:Type1}. We are not aware of any existing algorithms which use the sparsity measure induced by the BRST prior: $\sum_{g_b \in \mathscr{G}} \log(\Vert \vHgb \Vert_2^2+\tau)$.

\section{Analysis}
\label{section:analysis}
In this section, important properties of the proposed framework are analyzed. First, the properties of the framework as it applies to S-NNLS are studied. Then, the proposed framework is studied in the context of S-NMF and S-NMF-W.

\subsection{Analysis in the S-NNLS Setting}
\label{sec:convergence snnls}
We begin by confirming that \eqref{eq:Type 1 GEM M-step} does not have a trivial solution at $\Hij = \infty$ for any $(i,j)$ because $\left\langle \left(\gamma_{(i,j)}\right)^{-z}\right\rangle \geq 0$, since it is an expectation of a non-negative random variable. In the following discussion, it will be useful to work with distributions whose functional dependence on $\Hij$ has a power function form:
\begin{align}\label{eq:power function}
f(\Hij,z,\tau,\alpha) = \left(\tau+\myexp{\Hij}{z}\right)^{-\alpha}
\end{align}
where $\tau,\alpha > 0$ and $0 < z \leq 2$. Note that the priors considered in this work have a power function form.

\subsubsection{Monotonicity of $Q(H,\vHt)$ under \eqref{eq:mult rule nmf}}
The following theorem states one of the main contributions of this work, validating the use of \eqref{eq:mult rule nmf} in \eqref{eq:Type 1 GEM M-step}.
\begin{theorem}
\label{thm:nmf}
Let $z \in \lbrace 1,2\rbrace$ and the functional dependence of $\pr{\Hij}$ on $\Hij$ have a power function form. Consider using the update rule stated in \eqref{eq:mult rule nmf} to update $\vHijs{s}$ for all $(i,j) \in \mathscr{J}  = \lbrace (i,j) : \vHijs{s}>0 \rbrace$. Then, the update rule in \eqref{eq:mult rule nmf} is well defined and $Q(\vHs{s+1},\vHt) \leq Q(\vHs{s},\vHt)$.
\begin{proof}
Proof provided in \ref{appendix:proof nmf}.
\end{proof}
\end{theorem}
Theorem \ref{thm:nmf} also applies to the block-sparse MUR in \eqref{eq:mult rule block sparse nmf}.

\subsubsection{Local Minima of $L(H)$}
Before proceeding to the analysis of the convergence of Algorithm \ref{alg:Type 1}, it is important to consider the question as to whether the local minima of $L(H)$ are desirable solutions from the standpoint of being sparse.
\begin{theorem}\label{thm:local}
Let $\vHs{*}$ be a local minimum of \eqref{eq:Type1} and let the functional dependence of $\pr{\Hij}$ on $\Hij$ have a power function form. In addition, let one of the following conditions be satisfied: 1) $z \leq 1$ or 2) $z > 1$ and $\tau \rightarrow 0$. Then, $\left\Vert \vHjs{*} \right\Vert_0 \leq d$.
\begin{proof}
Proof provided in \ref{appendix:proof of local}.
\end{proof}
\end{theorem}

\subsubsection{Convergence of Algorithm \ref{alg:Type 1}}
First, an important property of the cost function in \eqref{eq:Type1} can be established.
\begin{theorem}\label{thm:rpesm coercive}
The function $-\log \p{\Hij}$ is coercive for any member of the RPESM family.
\begin{proof}
The proof is provided in \ref{appendix:proof of coercive theorem}.
\end{proof}
\end{theorem}
Theorem \ref{thm:rpesm coercive} can then be used to establish the following corollary.
\begin{corollary}\label{lemma:cost coercive}
Assume the signal model in \eqref{eq:signal model} and let $\p{\Hij}$ be a member of the RPESM family. Then, the cost function $L(H)$ in \eqref{eq:Type1} is coercive.
\begin{proof}
This follows from the fact that $\Vert X - WH \Vert_F^2 \geq 0$ and the fact that $-\log \p{\Hij}$ is coercive due to Theorem \ref{thm:rpesm coercive}.
\end{proof}
\end{corollary}
The coercive property of the cost function in \eqref{eq:Type1} allows us to establish the following result concerning Algorithm \ref{alg:Type 1}.
\begin{corollary}\label{lemma:existence of limit point}
Let $z \in \lbrace 1,2\rbrace$ and the functional dependence of $\pr{\Hij}$ on $\Hij$ have a power function form. Then, the sequence $\lbrace \bar{H}^t \rbrace_{t=1}^\infty$ produced by Algorithm \ref{alg:Type 1} with $S$, the number of inner loop iterations, set to $1$ admits at least one limit point.
\begin{proof}
The proof is provided in \ref{appendix:proof of existence of limit point lemma}.
\end{proof}
\end{corollary}

We are now in a position to state one of the main contributions of this paper regarding the convergence of Algorithm \ref{alg:Type 1} to the set of stationary points of \eqref{eq:Type1}. A stationary point is defined to be any point satisfying the Karush-Kuhn-Tucker (KKT) conditions for a given optimization problem \cite{boyd2004convex}.  

\begin{figure}
\centering
\includegraphics[scale=0.7]{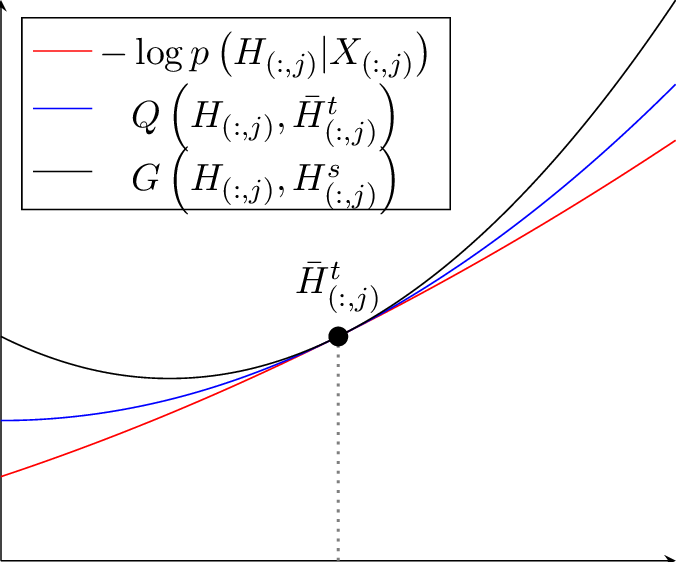}
\caption{Visualization of Algorithm \ref{alg:Type 1}}
\label{fig:EM visualization}
\end{figure}

\begin{theorem}\label{thm:global convergence}
Let $z \in \lbrace 1,2 \rbrace$, $\zeta = $ \eqref{eq:mult rule nmf}, $t^\infty = \infty$, $S = 1$, the functional dependence of $\pr{\Hij}$ on $\Hij$ have a power function form, the columns of $W$ and $X$ have bounded norm, and $W$ be full rank. In addition, let one of the following conditions be satisfied: (1) $z = 1 \text{ and } \tau \leq \lambda/\max_{i,j}\iji{W^T X}$ or (2) $z = 2 \text{ and } \tau \rightarrow 0$.
Then the sequence $\lbrace \bar{H}^t \rbrace_{t=1}^\infty$ produced by Algorithm \ref{alg:Type 1} is guaranteed to converge to the set of stationary points of $L(H)$. Moreover, $\lbrace L(\bar{H}^t)\rbrace_{t=1}^\infty$ converges monotonically to $L(\bar{H}^*)$, for stationary point $\bar{H}^*$.
\begin{proof}
The proof is provided in \ref{appendix:proof of stationary-fixed point}.
\end{proof}
\end{theorem}
The reason that $S=1$ is specified in Theorem \ref{thm:global convergence} is that it allows for providing convergence guarantees for Algorithm \ref{alg:Type 1} without needing any convergence properties of the sequence generated by \eqref{eq:mult rule nmf}. Theorem \ref{thm:global convergence} also applies to Algorithm \ref{alg:Type 1} when the block-sparse MUR in \eqref{eq:mult rule block sparse nmf} is used. To see the intuition behind the proof of Theorem \ref{thm:global convergence} (given in \ref{appendix:proof of stationary-fixed point}), consider the visualization of Algorithm \ref{alg:Type 1} shown in Fig. \ref{fig:EM visualization}. The proposed framework seeks a minimum of $-\log \p{\Hj | \vXj}$, for all $j$, through an iterative optimization procedure. At each iteration, $-\log \p{\Hj|\vXj}$ is bounded by the auxiliary function $\Q{\Hj,\vHjt{t}}$ \cite{boyd2004convex}\cite{dempster1977maximum}. This auxiliary function is then bounded by another auxiliary function, $G\left(\Hj,\vHjs{s}\right)$, defined in \eqref{eq:G}. Therefore, the proof proceeds by giving conditions under which \eqref{eq:Type 1 GEM M-step} is guaranteed to reach a stationary point of $-\log \p{\Hj|\vXj}$  by repeated minimization of $\Q{\Hj,\vHjt{t}}$ and then finding conditions under which $\Q{\Hj,\vHjt{t}}$ can be minimized by minimization of $G\left(\Hj,\vHjs{s}\right)$ through the use of \eqref{eq:mult rule nmf}.

\subsection{Analysis in S-NMF and S-NMF-W Settings}
We now extend the results of Section \ref{sec:convergence snnls} to the case where $W$ is unknown and is estimated using Algorithm \ref{alg:Sparse NMF}. For clarity, let $(z_w,\tau_w)$ and $(z_h,\tau_h)$ refer to the distributional parameters of the priors over $\vect{W}$ and $\vH$, respectively. As before, $\tau_w,\tau_h > 0$ and $0 < z_w,z_h \leq 2$. First, it is confirmed that Algorithm \ref{alg:Sparse NMF} exhibits the same desirable optimization properties as the NMF MUR's \eqref{eq:MUR W}-\eqref{eq:MUR H}.

\begin{corollary}
\label{cor:1}
Let $z_w,z_h \in \lbrace 1,2 \rbrace$ and the functional dependence of $\pr{\Hij}$ on $\Hij$ have a power function form. If performing S-NMF-W, let the functional dependence of $\pr{\Wij}$ on $\Wij$ have a power function form. Consider using Algorithm \ref{alg:Sparse NMF} to generate $\left\lbrace\bar{W}^t,\bar{H}^t\right\rbrace_{t=0}^\infty$. Then, the update rules used in Algorithm \ref{alg:Sparse NMF} are well defined and $L^{NMF}\left(\bar{W}^{t+1},\bar{H}^{t+1}\right) \leq L^{NMF}\left(\bar{W}^t,\bar{H}^t\right)$.
\begin{proof}
The proof is shown in \ref{appendix:proof of cor 1}.
\end{proof}
\end{corollary}
Therefore, the proposed S-NMF framework maintains the monotonicity property of the original NMF MUR's, with the added benefit of promoting sparsity in $H$ (and $W$, in the case of S-NMF-W). Unfortunately, it is not clear how to obtain a result like Theorem \ref{thm:global convergence} for Algorithm \ref{alg:Sparse NMF} in the S-NMF setting. The reason that such a result cannot be shown is because it is not clear that if a limit point, $(\bar{W}^{\infty},\bar{H}^{\infty})$, of Algorithm \ref{alg:Sparse NMF} exists, that this point is a stationary point of $L^{NMF}(\cdot,\cdot)$. Specifically, if there exists $(i,j)$ such that $\bar{W}_{(i,j)}^{\infty} = 0$, the KKT condition $-(X-\bar{W}^{\infty} \bar{H}^\infty) \left(\bar{H}^{\infty}\right)^T \geq 0$ cannot be readily verified. This deficiency is unrelated to the size of $W$ and $H$ and is, in fact, the reason that convergence guarantees for the original update rules in \eqref{eq:MUR W}-\eqref{eq:MUR H} do not exist. Interestingly, if Algorithm \ref{alg:Sparse NMF} is considered in S-NMF-W mode, this difficulty is alleviated.

\begin{corollary}\label{lemma:existence of limit point nmf}
Let $z_w,z_h \in \lbrace 1,2\rbrace$, $S = 1$, and the functional dependence of $\pr{\Hij}$ on $\Hij$ and of $\pr{\Wij}$ on $\Wij$ have power function forms. Then, the sequence $\lbrace \bar{H}^t,\bar{W}^t \rbrace_{t=1}^\infty$ produced by Algorithm \ref{alg:Sparse NMF} admits at least one limit point.
\begin{proof}
The objective function is now coercive with respect to $W$ and $H$ as a result of the application of Theorem \ref{thm:rpesm coercive} to $-\log \pr{\Hij}$ and $-\log \pr{\Wij}$. Since $\lbrace L^{NMF}(\bar{W}^t,\bar{H}^t) \rbrace_{t=1}^\infty$ is a non-increasing sequence, the proof for Corollary \ref{lemma:existence of limit point} in \ref{appendix:proof of existence of limit point lemma} can be applied to obtain the stated result. 
\end{proof}
\end{corollary}

\begin{corollary}\label{thm:global convergence nmf}
Let $\lbrace\bar{W}^{t},\bar{H}^{t}\rbrace_{t=1}^\infty$ be a sequence generated by Algorithm \ref{alg:Sparse NMF} with $\zeta_w = $ \eqref{eq:mult rule nmf}. Let $z_h,z_w \in \lbrace 1,2\rbrace$, the functional dependence of $\pr{\Hij}$ on $\Hij$ have a power function form, the functional dependence of $\pr{\Wij}$ on $\Wij$ have a power function form, the columns and rows of $X$ have bounded norm, the columns of $\bar{W}^\infty$ have bounded norm, the rows of $\bar{H}^\infty$ have bounded norm, and $\bar{W}^\infty$ and $\bar{H}^\infty$ be full rank. Let one of the following conditions be satisfied: (1h) $z_h = 1 \text{ and } \tau_h \leq \lambda/\max_{i,j}\iji{\left(\bar{W}^\infty\right)^T X}$ or (2h)
$z_h = 2 \text{ and } \tau_h \rightarrow 0$.
In addition, let one of the following conditions be satisfied: (1w)
$z_w = 1 \text{ and } \tau_w \leq \lambda/{\max_{i,j}\iji{\bar{H}^\infty X^T}}$
or (2w)
$z_w = 2 \text{ and } \tau_w \rightarrow 0$.
Then, $\lbrace\bar{W}^{t},\bar{H}^{t}\rbrace_{t=1}^\infty$ is guaranteed to converge to set of stationary points of $L^{NMF}(\cdot,\cdot)$.
\begin{proof}
The proof is provided in \ref{appendix:proof of thm 4}.
\end{proof}
\end{corollary}

\begin{figure*}
\begin{subfigure}{0.5\textwidth}
\centering\includegraphics[width=\textwidth]{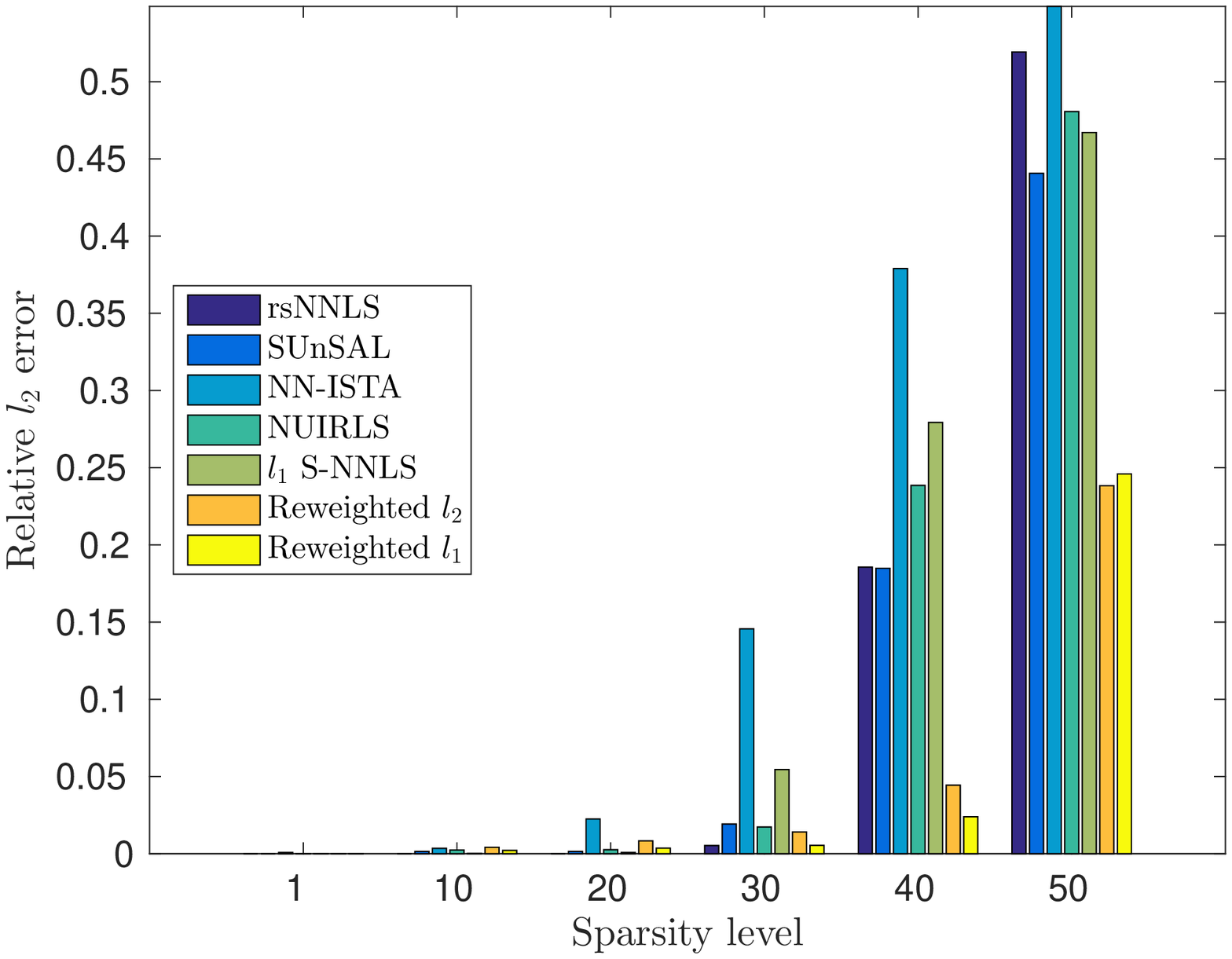}
\caption{Average relative $\ell_2$ error as a function of sparsity level for $n = 400$.}
\label{fig:sparse recovery as function of s}
\end{subfigure}
~
\begin{subfigure}{0.5\textwidth}
\centering\includegraphics[width=\textwidth]{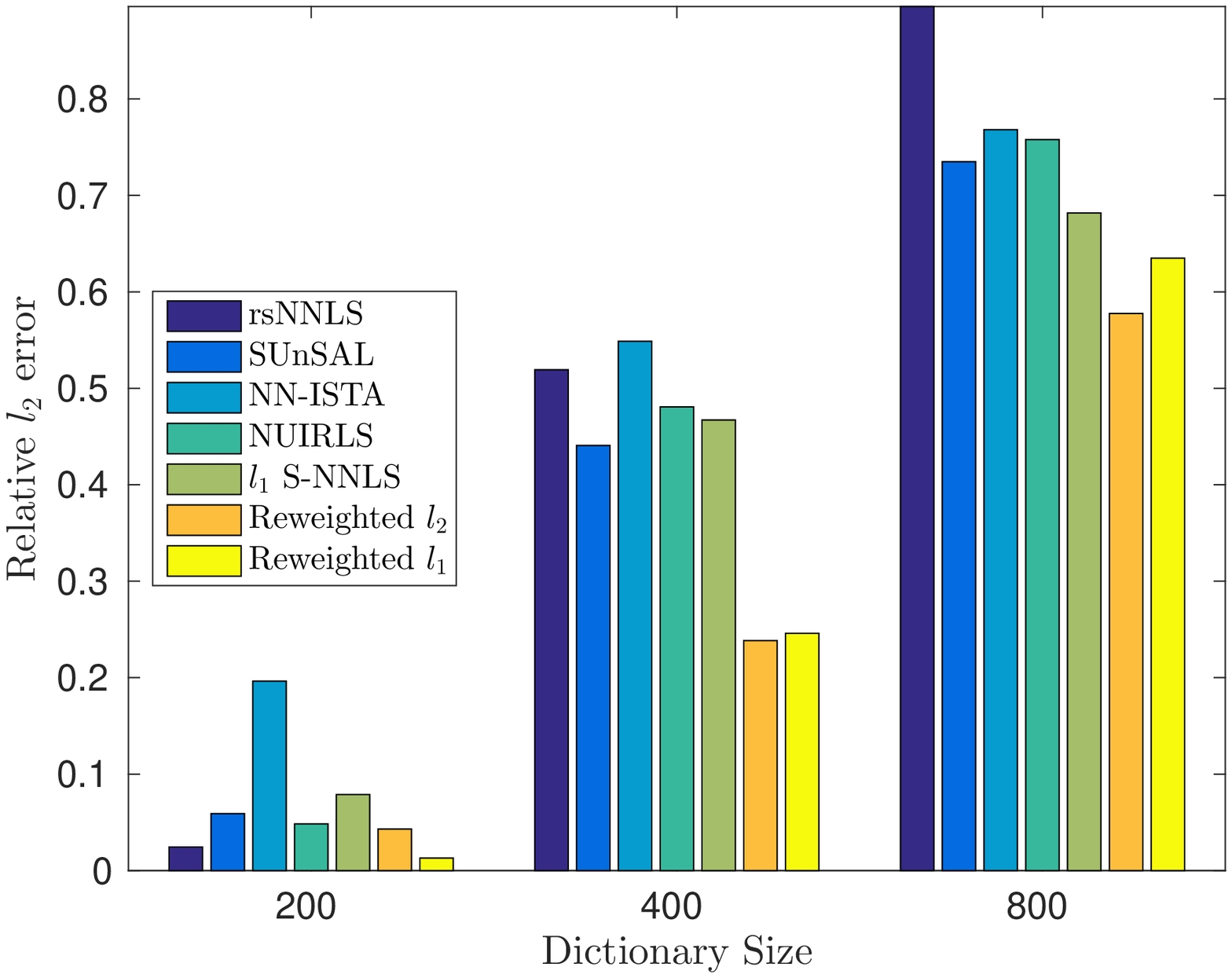}
\caption{Average relative $\ell_2$ error as a function of $n$ for sparsity level $50$.}
\label{fig:sparse recovery as function of d}
\end{subfigure}

\begin{subfigure}{0.5\textwidth}
\centering\includegraphics[width=\textwidth]{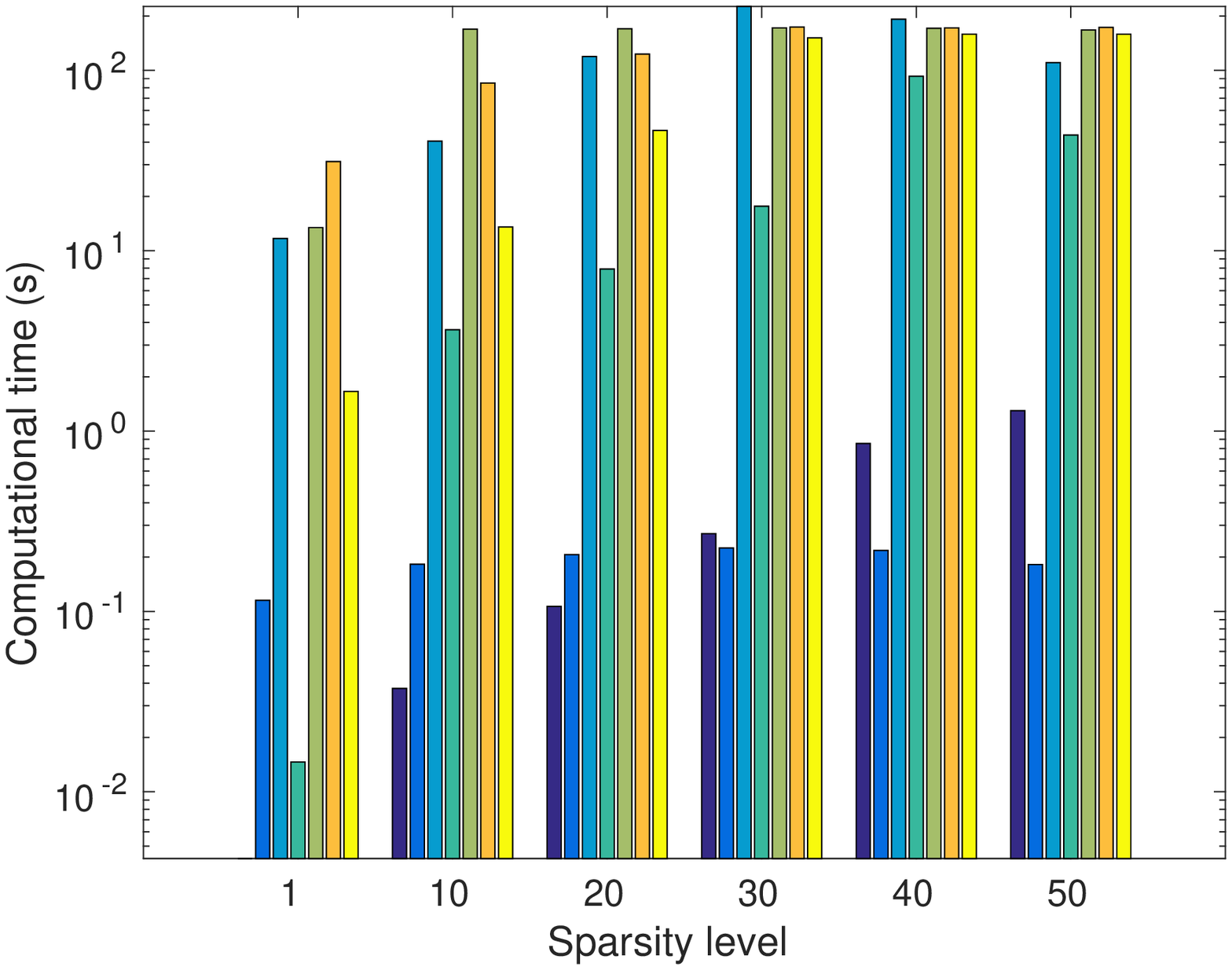}
\caption{Average computational time as a function of sparsity level for $n = 400$.}
\label{fig:computational time as function of s}
\end{subfigure}
~
\begin{subfigure}{0.5\textwidth}
\centering\includegraphics[width=\textwidth]{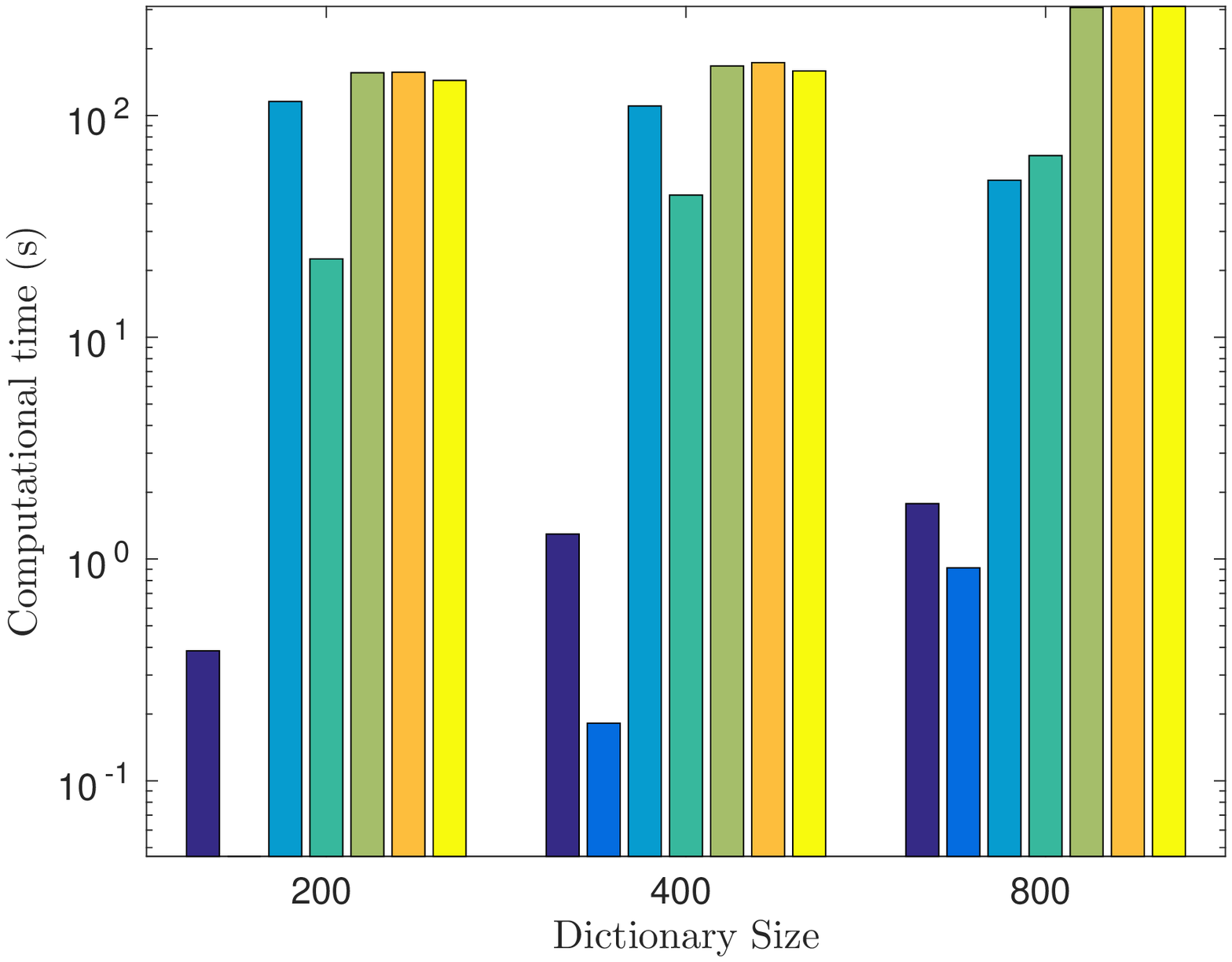}
\caption{Average computational time as a function of $n$ for sparsity level $50$.}
\label{fig:computational time as function of d}
\end{subfigure}
\caption{S-NNLS results on synthetic data. The legends for (c) and (d) have been omitted, but are identical to the legends in (a) and (b).}
\label{fig:matrix sparse coding results}
\end{figure*}

\section{Experimental Results}
\label{section:results}
In the following, experimental results for the class of proposed algorithms are presented. The experiments performed were designed to highlight the main properties of the proposed approaches. First, the accuracy of the proposed S-NNLS algorithms on synthetic data is studied. Then, experimental validation for claims made in Section \ref{section:analysis} regarding the properties of the proposed approaches is provided. Finally, the proposed framework is shown in action on real-world data by learning a basis for a database of face images. 

\subsection{S-NNLS Results on Synthetic Data}
\label{section:synthetic results}
In order to compare the described methods, a sparse recovery experiment was undertaken. First, a dictionary $W \in \mathbb{R}^{100 \times n}_+$ is generated, where each element of $W$ is drawn from the $\mathsf{RG}(0,1)$ distribution. The columns of $W$ are then normalized to have unit $\ell_2$ norm. The matrix $H \in \mathbb{R}_+^{n \times 100}$ is then generated by randomly selecting $k$ coefficients of $\Hj$ to be non-zero and drawing the non-zero values from a $\mathsf{RG}(0,1)$ distribution. The columns of $H$ are normalized to have unit $\ell_2$ norm. We then feed $X = W H$ and $W$ to the S-NNLS algorithm and approximate $\Hj$ with ${\hat{H}_{(:,j)}}$. Note that this is a noiseless experiment. The distortion of the approximation is measured using the relative Frobenius norm error, ${\Vert H - \hat{H} \Vert_F}/{\Vert H \Vert_F}$. A total of $50$ trials are run and averaged results are reported. 

We use Algorithm \ref{alg:Type 1} to generate recovery results for the proposed framework, with the number of inner-loop iterations, $S$, of Algorithm \ref{alg:Type 1} set to $2000$ and the outer EM loop modified to run a maximum of $50$ iterations. For reweighted $\ell_2$ S-NNLS, the same annealing strategy for $\tau$ as reported in \cite{chartrand2008iteratively} is employed, where $\tau$ is initialized to $1$ and decreased by a factor of $10$ (up to a pre-specified number of times) when the relative $\ell_2$ difference between $\vHjt{t+1}$ and $\vHjt{t}$ is below ${\sqrt{\tau}}/{100}$ for each $j$. Note that this strategy does not influence the convergence properties described in Section \ref{section:analysis} for the reweighted $\ell_2$ approach since $\tau$ can be viewed as fixed after a certain number of iterations. For reweighted $\ell_1$ S-NNLS, we use $\tau = 0.1$. The regularization parameter $\lambda$ is selected using cross-validation by running the S-NNLS algorithms on data generated using the same procedure as the test data. 

We compare our results with rsNNLS \cite{peharz2012sparse}, the SUnSAL algorithm for solving \eqref{eq:l1 NMF} \cite{bioucas2010alternating}, the non-negative ISTA (NN-ISTA) algorithm\footnote{We modify the soft-thresholding operator to $S_\beta(h) = max(0,\vert h \vert -\beta)$} for solving \eqref{eq:l1 NMF} \cite{daubechies2004iterative}, NUIRLS, and $\ell_1$ S-NNLS \cite{hoyer2004non} (i.e \eqref{eq:MUR H l1}). Since rsNNLS requires $k$ as an input, we incorporate knowledge of $k$ into the tested algorithms in order to have a fair comparison. This is done by first thresholding ${\hat{H}_{(:,j)}}$ by zeroing out all of the elements except the largest $k$ and then executing \eqref{eq:MUR H} until convergence. 

The S-NNLS results are shown in Fig. \ref{fig:matrix sparse coding results}. Fig. \ref{fig:sparse recovery as function of s} shows the recovery results for $n = 400$ as a function of the sparsity level $k$. All of the tested algorithms perform almost equally well up to $k = 30$, but the reweighted approaches dramatically outperform the competing methods for $k = 40$ and $k = 50$. Fig. \ref{fig:sparse recovery as function of d} shows the recovery results for $k = 50$ as a function of $n$. All of the tested algorithms perform relatively well for $n = 200$, but the reweighted approaches separate themselves  for $n = 400$ and $n = 800$. Fig. \ref{fig:computational time as function of s} and \ref{fig:computational time as function of d} show the average computational time for the algorithms tested as a function of sparsity level and dictionary size, respectively.

Two additional observations from the results in Fig. \ref{fig:sparse recovery as function of s} can be made. First, the reweighted approaches perform slightly worse for sparsity levels $k \leq 20$. We believe that this is a result of suboptimal parameter selection for the reweighted algorithms and using a finer grid during cross-validation would improve the result. This claim is supported by the observation that NUIRLS performs at least as well or better than the reweighted approaches for $k \leq 20$ and, as argued in Section \ref{section:S-NNLS rl2}, NUIRLS is equivalent to reweighted $\ell_2$ S-NNLS in the limit $\lambda,\tau \rightarrow 0$. The second observation is that the reweighted $\ell_2$ approach consistently outperforms NUIRLS at high values of $k$. This suggests that the strategy of allowing $\lambda > 0$ and annealing $\tau$, instead of setting it to $0$ as in NUIRLS \cite{grady2008compressive}, is much more robust.

\begin{figure*}
\centering
\begin{minipage}{0.45\textwidth}
\centering
\includegraphics[width=0.9\textwidth]{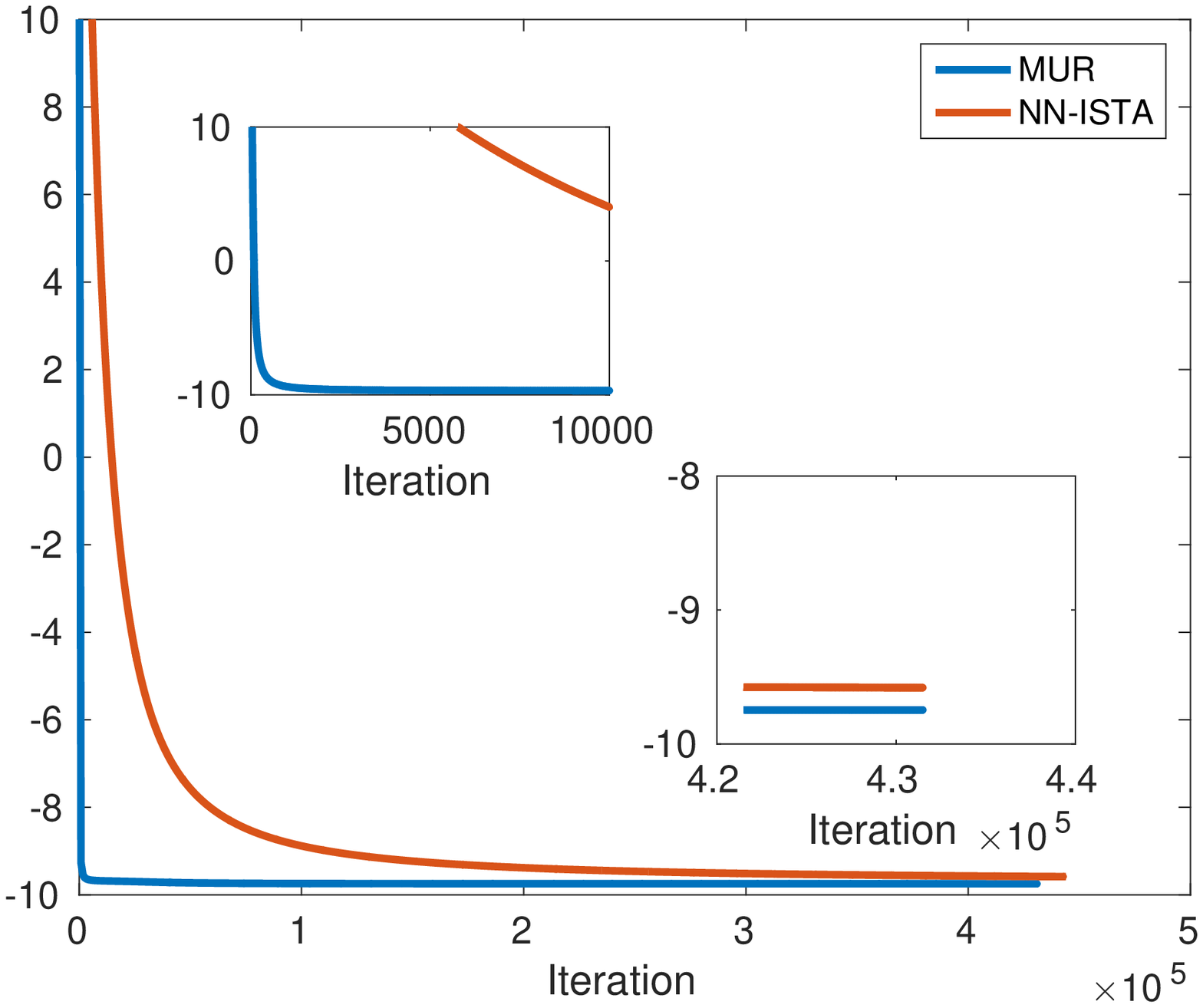} 
\caption{Evolution of $L(H)$ for the reweighted $\ell_1$ formulation in Section \ref{section: S-NNLS rl1} using Algorithm \ref{alg:Type 1} and a baseline approach employing the NN-ISTA algorithm.}
\label{fig:comparison}
\end{minipage}\hfill
\begin{minipage}{0.45\textwidth}
\centering
\includegraphics[width=0.9\textwidth]{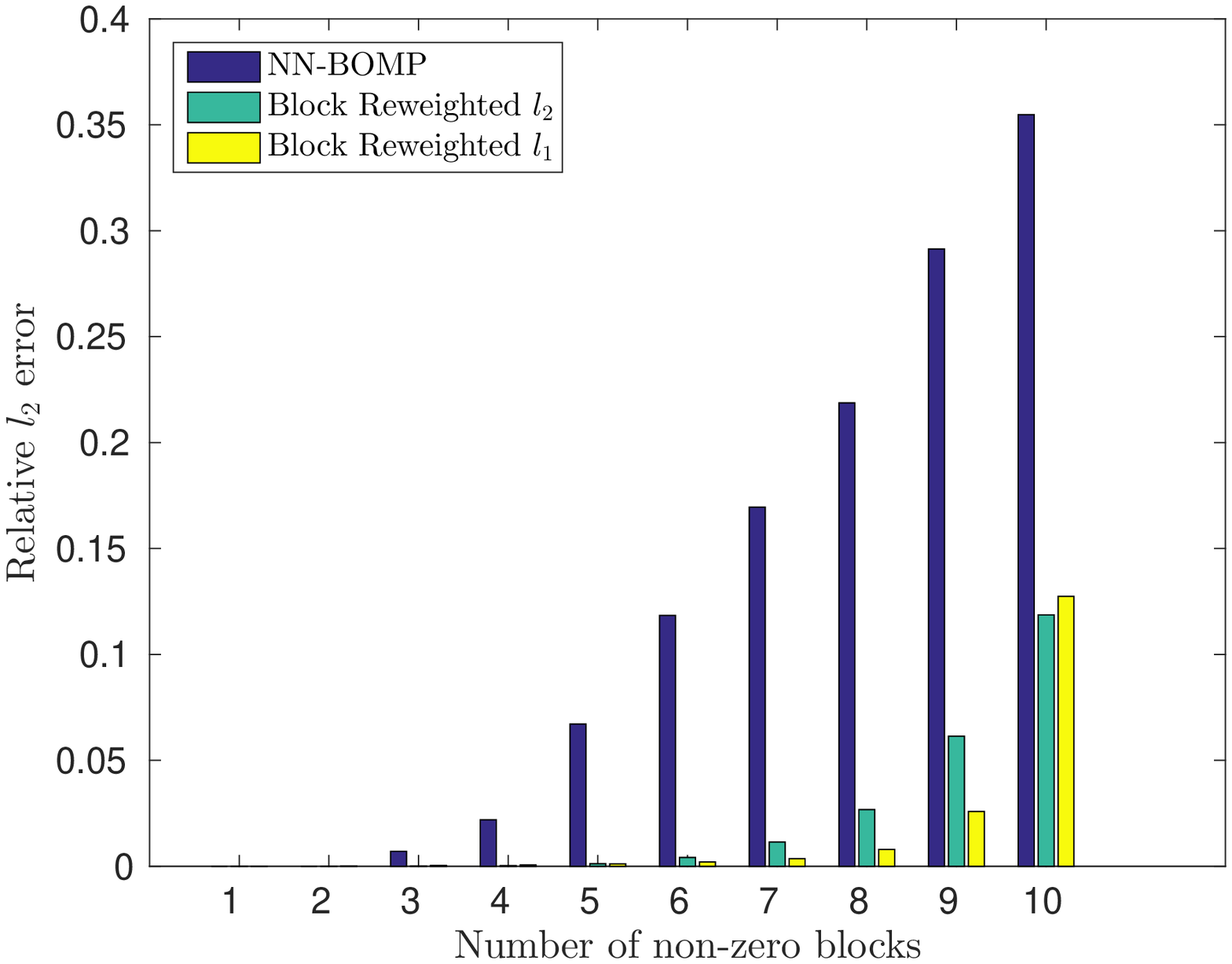}
\caption{Block sparse recovery results}
\label{fig:block sparse}
\end{minipage}
\end{figure*}

In addition to displaying superior S-NNLS performance, the proposed class of MUR's also exhibits fast convergence. Fig. \ref{fig:comparison} compares the evolution of the objective function $L(H)$ under the RGDP signal prior (i.e. the reweighted $\ell_1$ formulation of Section \ref{section: S-NNLS rl1}) for Algorithm \ref{alg:Type 1}, with $S=1$, with a baseline approach. The baseline employs the NN-ISTA algorithm to solve the reweighted $\ell_1$ optimization problem which results from bounding the regularization term by a linear function of $H_{(i,j)}$ (similar to \eqref{eq:rl2 nuirls}, but with $\Vert H/Q^t \Vert_F^2$ replaced by $\Vert H/Q^t \Vert_1$). The experimental results show that the MUR in \eqref{eq:mult rule rl1} achieves much faster convergence as well as a lower objective function value compared to the baseline.

\subsection{Block S-NNLS Results on Synthetic Data}
In this experiment, we first generate $W \in \mathbb{R}_+^{80 \times 160}$ by drawing its elements from a $\mathsf{RG}(0,1)$ distribution. We generate the columns of $H \in \mathbb{R}_+^{160 \times 100}$ by partitioning each column into blocks of size $8$ and randomly selecting $k$ blocks to be non-zero. The non-zero blocks are filled with elements drawn from a $\mathsf{RG}(0,1)$ distribution. We then attempt to recover $H$ from $X = W H$. The relative Frobenius norm error is used as the distortion metric and results averaged over $50$ trials are reported.

The results are shown in Fig. \ref{fig:block sparse}. We compare the greedy NN-BOMP algorithm with the reweighted approaches. The reweighted approaches consistently outperform the $\ell_0$ based method, showing good recovery performance even when the number of non-zero elements of each column of $H$ is equal to the dimensionality of the column.


\subsection{A Numerical Study of the Properties of the Proposed Methods}
\begin{figure*}
\begin{subfigure}{0.3\textwidth}
\centering\includegraphics[width=\textwidth]{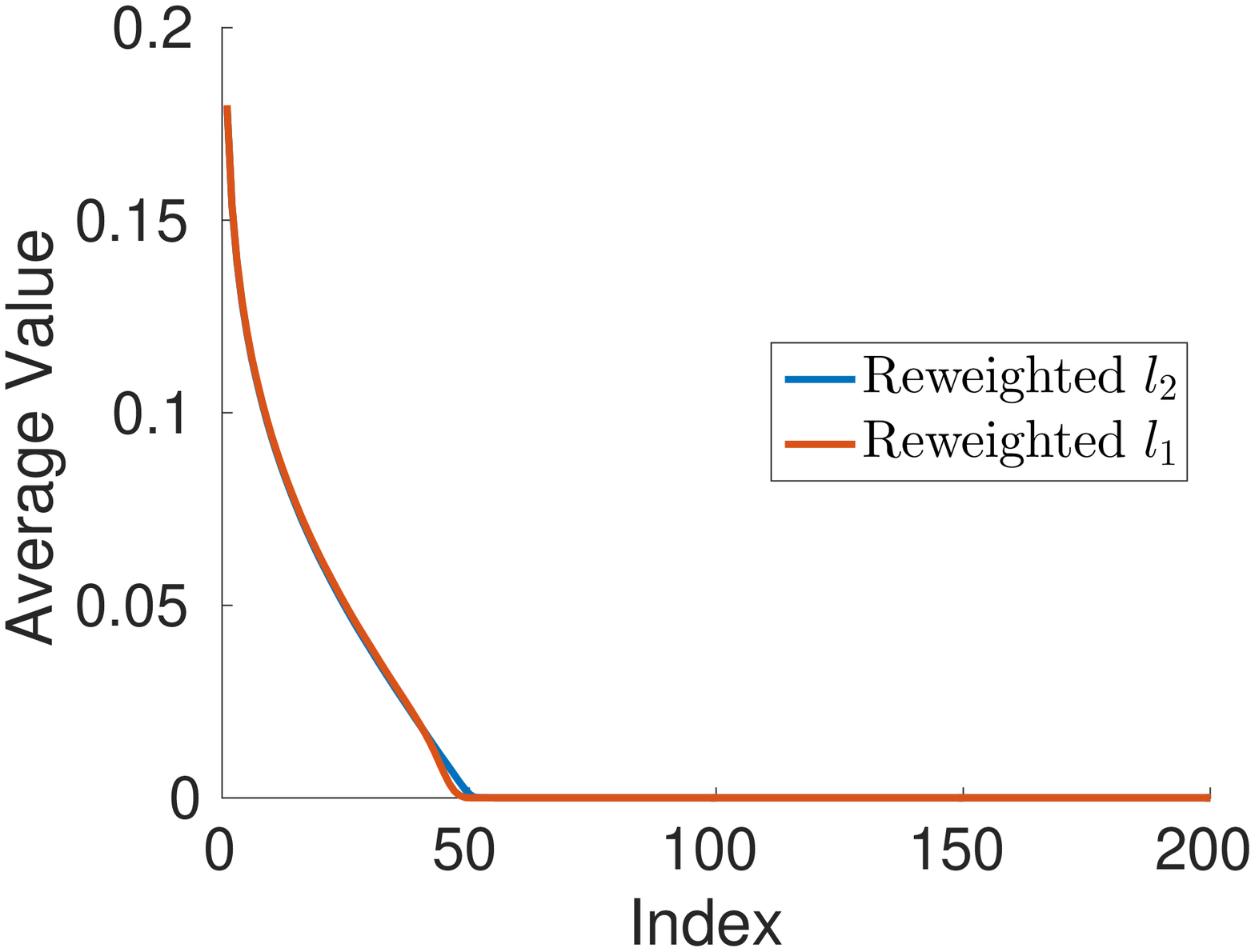}
\caption{$n = 200$}
\label{subfig:basic a}
\end{subfigure}
~
\begin{subfigure}{0.3\textwidth}
\centering\includegraphics[width=\textwidth]{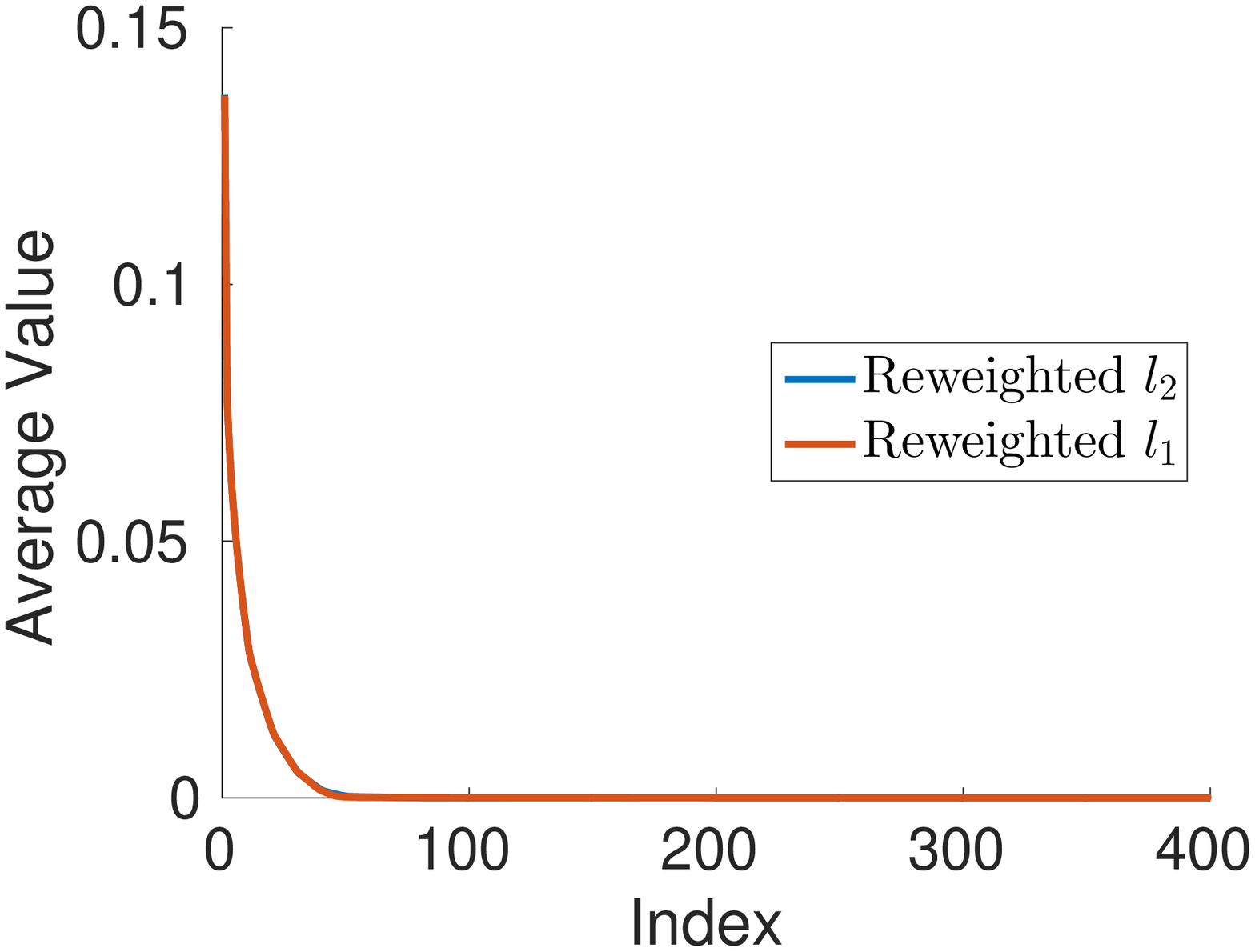}
\caption{$n = 400$}
\label{subfig:basic b}
\end{subfigure}
~
\begin{subfigure}{0.3\textwidth}
\centering\includegraphics[width=\textwidth]{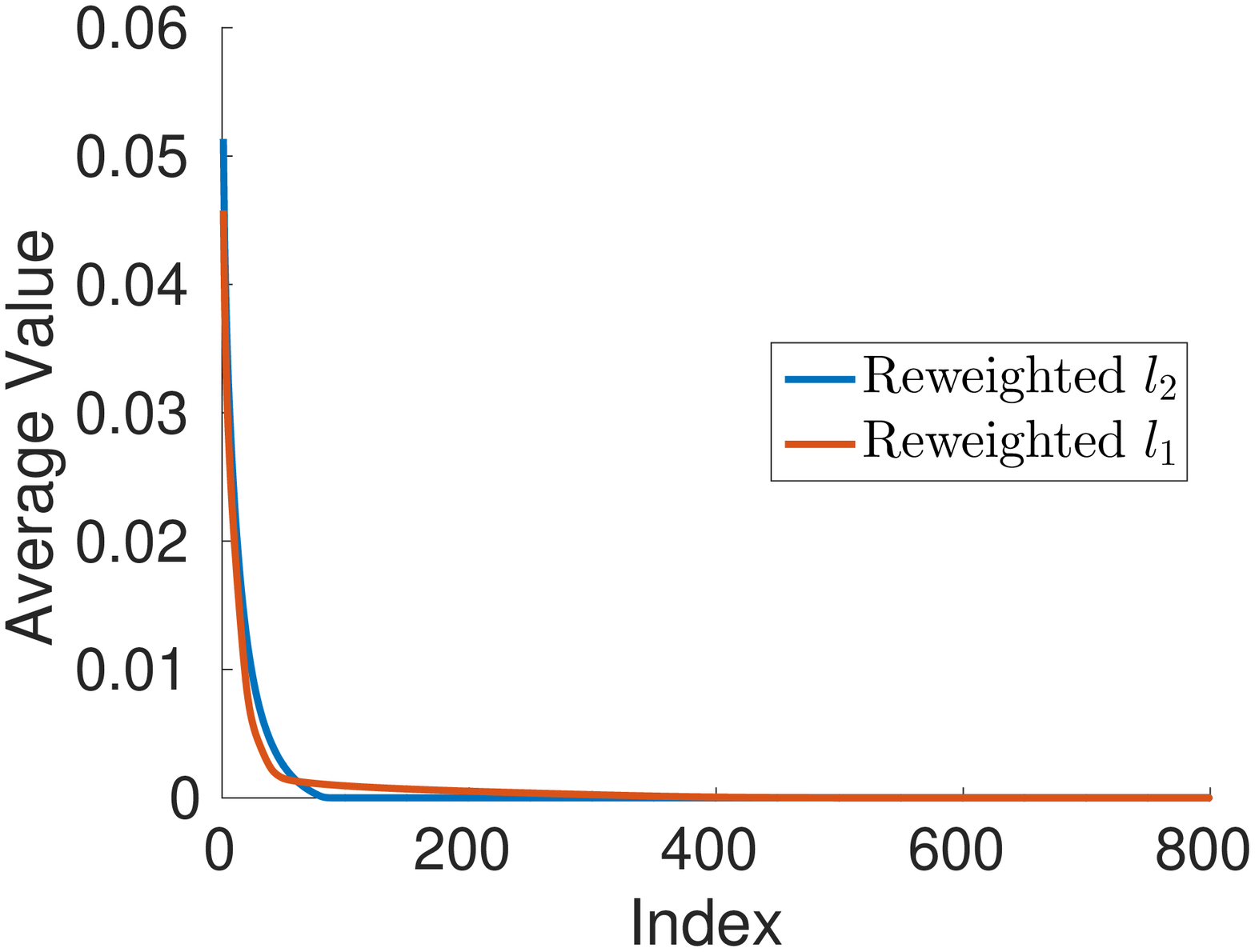}
\caption{$n = 800$}
\label{subfig:basic c}
\end{subfigure}

\caption{Average sorted coefficient value for S-NNLS with $W = \mathbb{R}_+^{100 \times n}$. The value at index $n_0$ represents the average value of the $n_0$'th largest coefficient the esitmated $\hat{H}$.}
\label{fig:sorted coefficients}
\end{figure*}

In this section, we seek to provide experimental verification for the claims made in the Section \ref{section:analysis}. First, the sparsity of the solutions obtained for the synthetic data experiments described in Section \ref{section:synthetic results} is studied. Fig. \ref{fig:sorted coefficients} shows the magnitude of the $n_0$'th largest coefficient in $\hat{H}_{(:,j)}$ for various sizes of $W$, averaged over all $50$ trials, all $j$, and all sparsity levels tested. The statement in Theorem \ref{thm:local} claims that the local minima of the objective function being optimized are sparse (i.e. that the number of nonzero entries is at most $d = 100$). In general, the proposed methods cannot be guaranteed to converge to a local minimum as opposed to a saddle point, so it cannot be expected that every solution produced by Algorithm \ref{alg:Type 1} is sparse. Nevertheless, Fig. \ref{fig:sorted coefficients} shows that for $n = 200$ and $n =400$, both reweighted approaches consistently find solutions with sparsity levels much smaller than $100$. For $n = 800$, the reweighted $\ell_2$ approach still finds solutions with sparsity smaller than $100$, but the reweigthed $\ell_1$ method deviates slightly from the general trend.


\begin{table}
\parbox{.45\linewidth}{
\centering
\resizebox{0.45\columnwidth}{!}{
\begin{tabular}{| c | c | c |c|}
\hline
$n$ & $200$ & $400$ & $800$\\  \hline
Reweighted $\ell_2$ & $10^{-9.3}$ & $10^{-9.4}$ & $10^{-9.6}$ \\
Reweighted $\ell_1$ &  $10^{-9.9}$ & $10^{-10.1}$ & $10^{-10.4}$\\ \hline
\end{tabular}}
\caption{Normalized KKT residual for S-NNLS algorithms on synthetic data. For all experiments, $d = 100$ and $k = 10$.}
\label{table:snnls kkt}
}
\hfill
\parbox{.45\linewidth}{
\centering
\resizebox{0.45\columnwidth}{!}{
\begin{tabular}{| c | c | c |}
\hline
 & $W$ & $H$\\  \hline
Reweighted $\ell_2$ & $10^{-3.9}$ & $10^{-5.3}$\\
Reweighted $\ell_1$ &  $10^{-5}$ & $10^{-7.3}$ \\ \hline
\end{tabular}}
\caption{Normalized KKT residual for S-NMF-W algorithms on CSBL face dataset.}
\label{table:nmf kkt}
}
\end{table}

Next, we test the claim made in Theorem \ref{thm:global convergence} that the proposed approaches reach a stationary point of the objective function by monitoring the KKT residual norm of the scaled objective function. Note that, as in \ref{appendix:proof of stationary-fixed point}, the $-\log u(\Hij)$ terms are omitted from $L(H)$ and the minimization of $L(H)$ is treated as a constrained optimization problem when deriving KKT conditions. For instance, for reweighted $\ell_1$ S-NNLS, the KKT conditions can be stated as
\begin{align}
\min\left(H,W^T WH - W^T X + \lambda\frac{\tau+1}{\tau + H}\right) = 0
\end{align}
and the norm of the left-hand side, averaged over all of the elements of $H$, can be viewed as a measure of how close a given $H$ is to being stationary \cite{gonzalez2005accelerating}. Table \ref{table:snnls kkt} shows the average KKT residual norm of the scaled objective function for the reweighted approaches for various problem sizes. The reported values are very small and provide experimental support for Theorem \ref{thm:global convergence}.

\begin{figure*}
\begin{subfigure}{0.3\textwidth}
\centering\includegraphics[width=\textwidth]{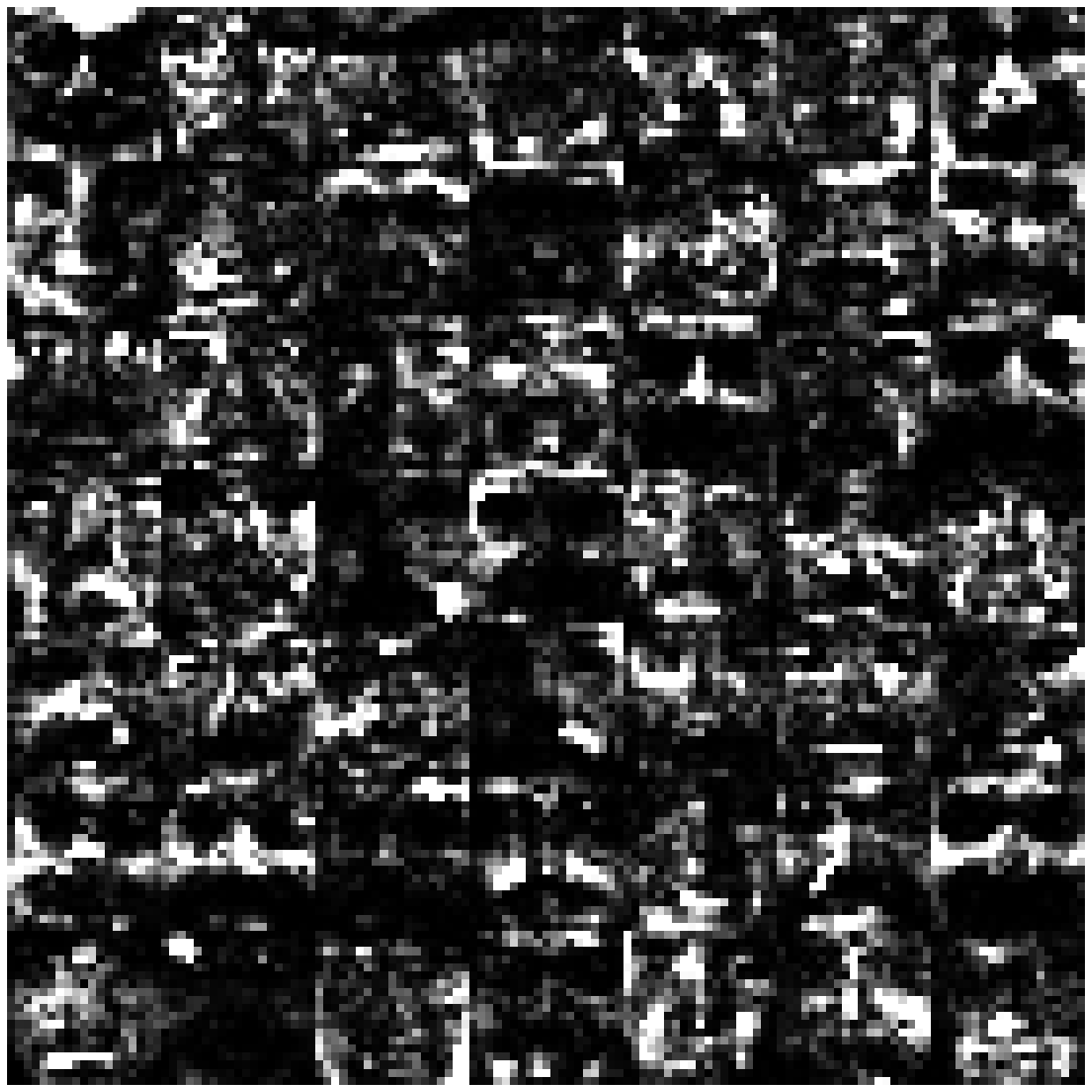}
\caption{}
\label{subfig:1}
\end{subfigure}
~
\begin{subfigure}{0.3\textwidth}
\centering\includegraphics[width=\textwidth]{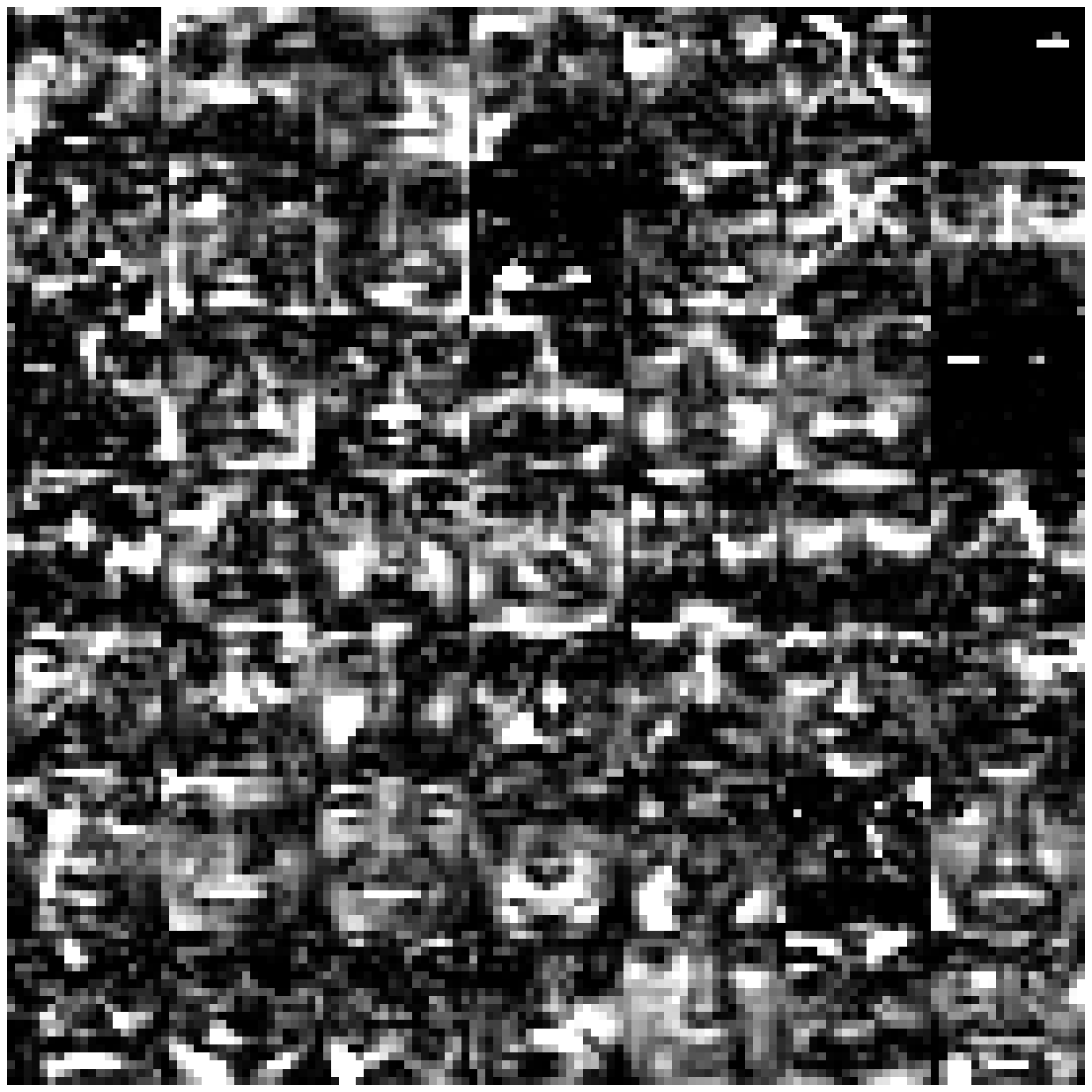}
\caption{}
\label{subfig:2}
\end{subfigure}
~
\begin{subfigure}{0.3\textwidth}
\centering\includegraphics[width=\textwidth]{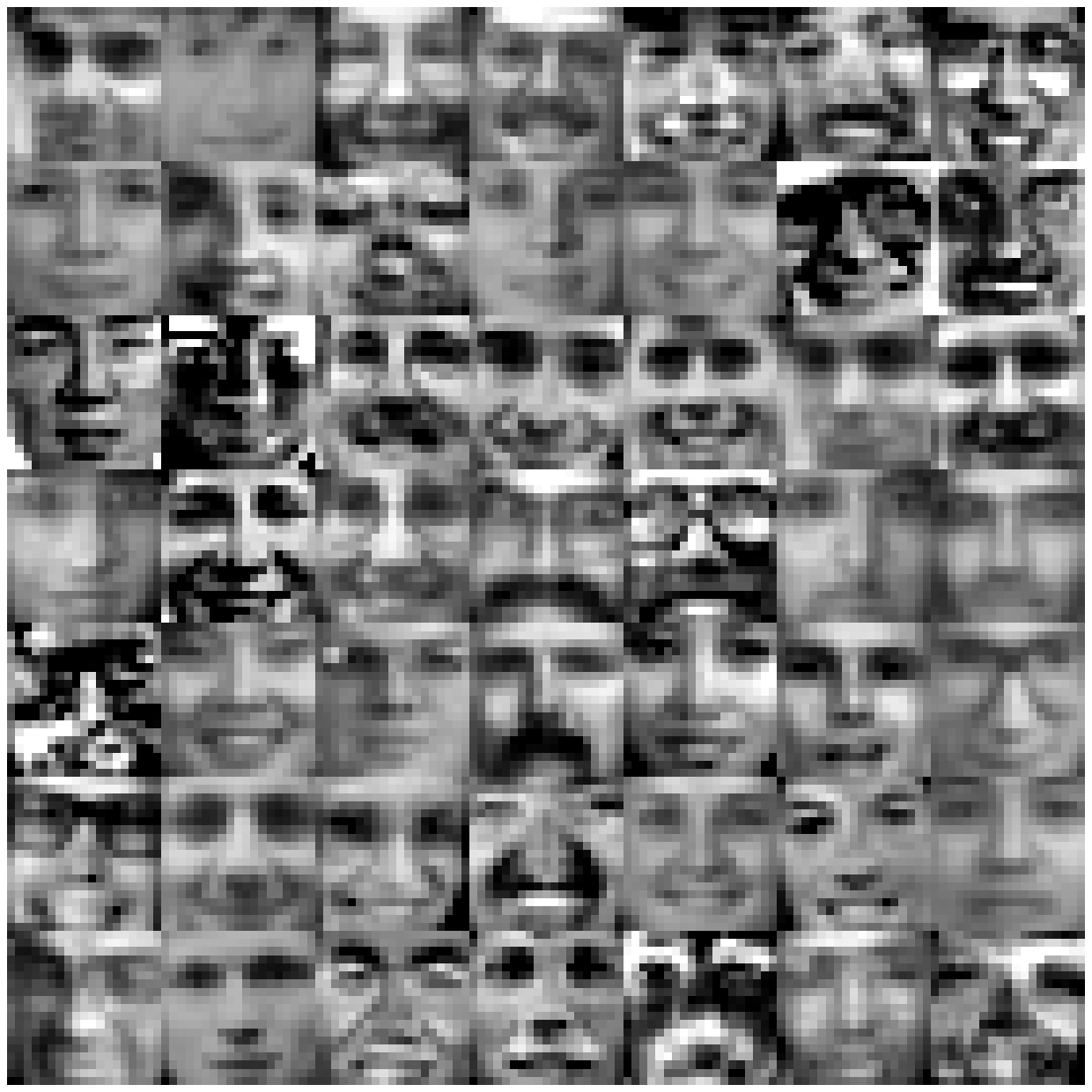}
\caption{}
\label{subfig:3}
\end{subfigure}

\begin{subfigure}{0.3\textwidth}
\centering\includegraphics[width=\textwidth]{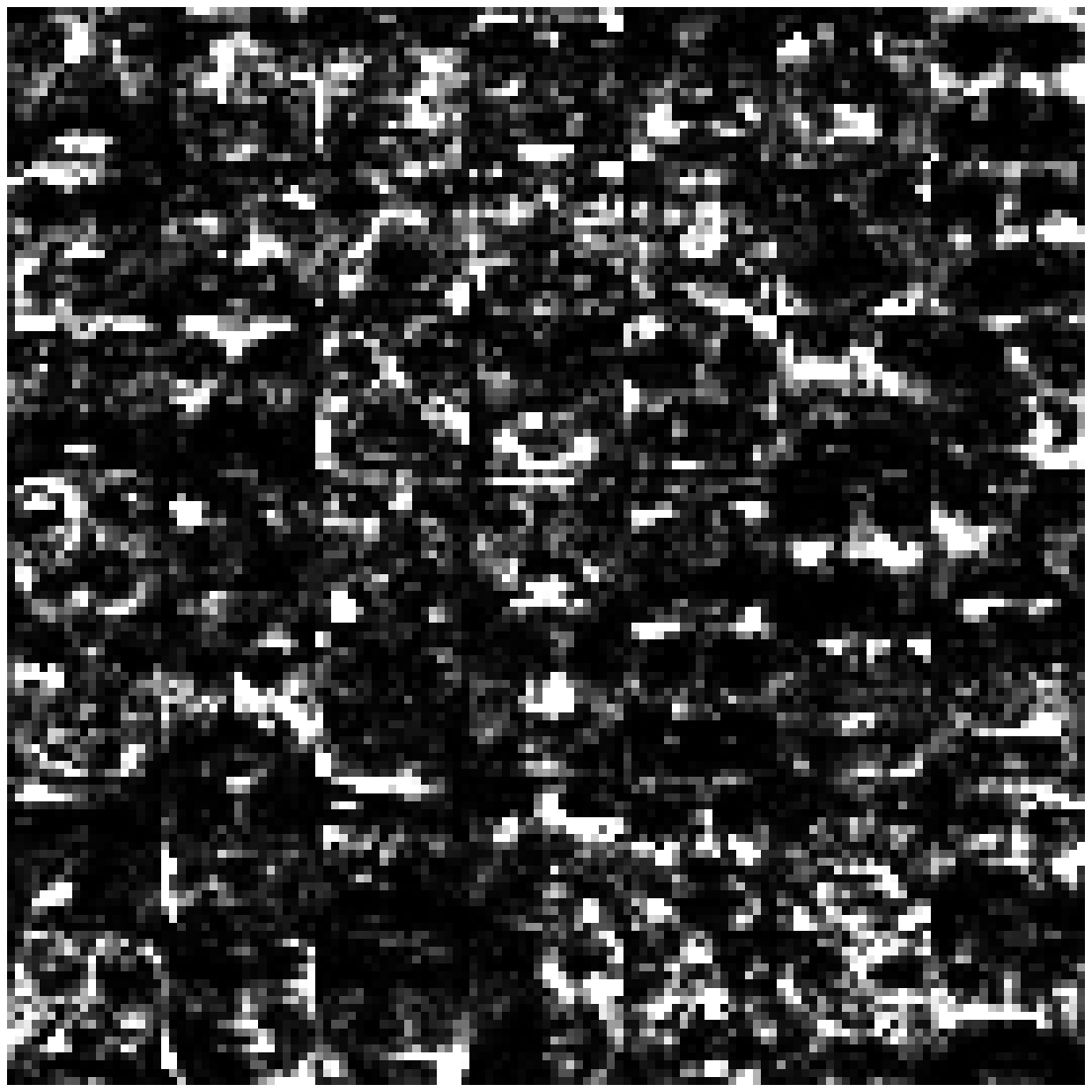}
\caption{}
\label{subfig:4}
\end{subfigure}
~
\begin{subfigure}{0.3\textwidth}
\centering\includegraphics[width=\textwidth]{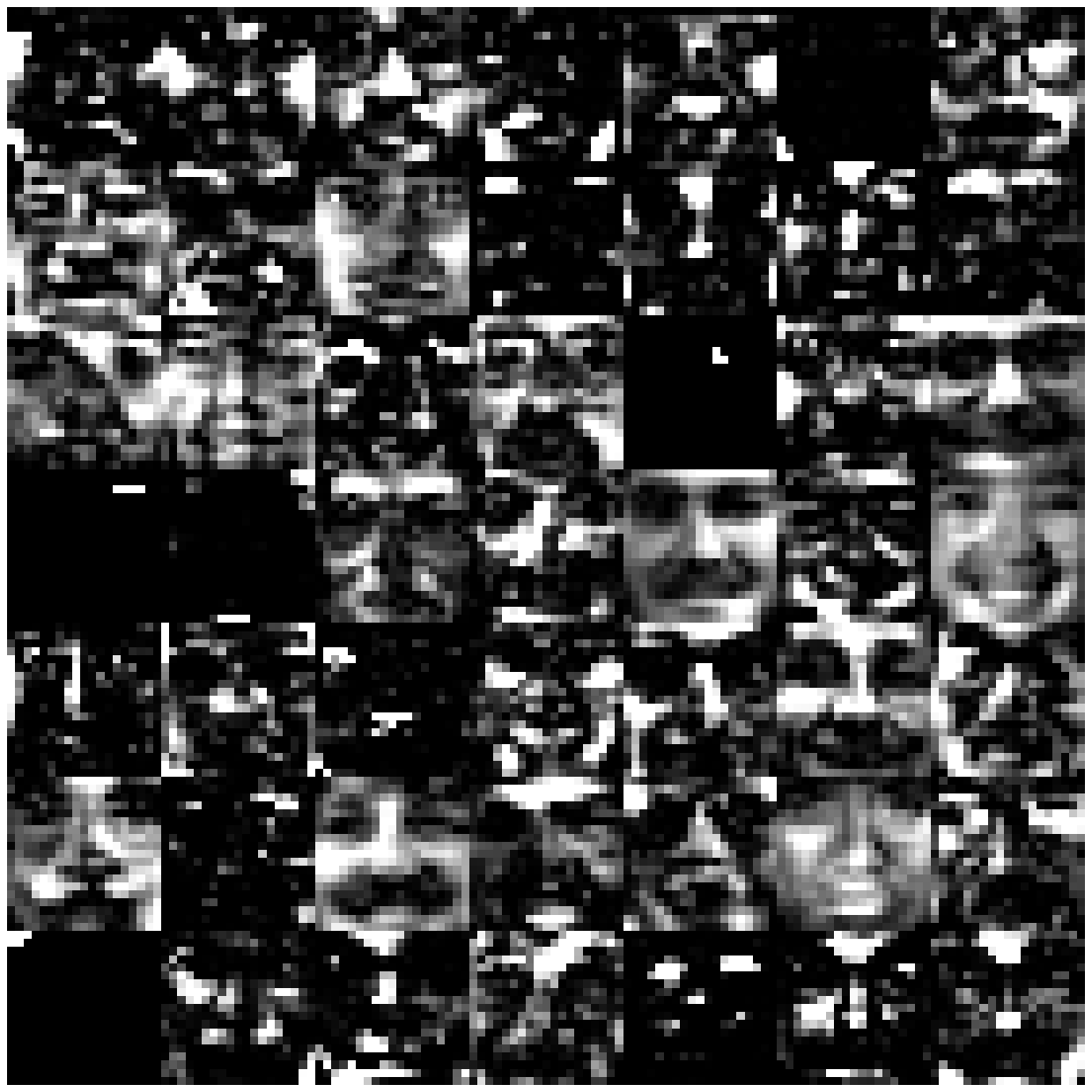}
\caption{}
\label{subfig:5}
\end{subfigure}
~
\begin{subfigure}{0.3\textwidth}
\centering\includegraphics[width=\textwidth]{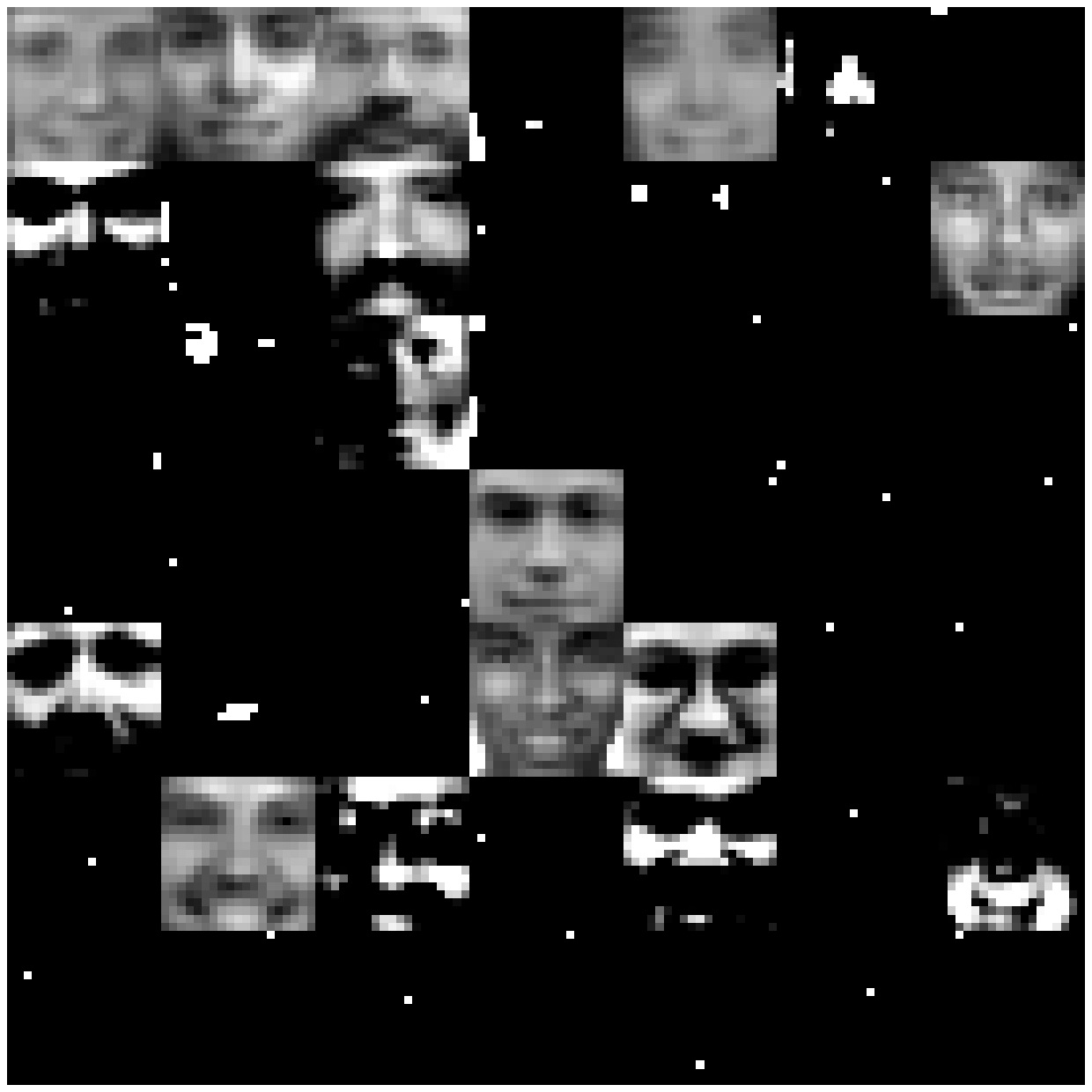}
\caption{}
\label{subfig:6}
\end{subfigure}

\caption{Visualization of random subset of learned atoms of $W$ for CBCL dataset. \ref{subfig:1}-\ref{subfig:3}: S-NMF with reweighted $\ell_1$ regularization on $H$, $\lambda = 1e-3, 1e-2, 1e-1$, respectively. \ref{subfig:4}-\ref{subfig:6}: S-NMF-W with reweighted $\ell_1$ regularization on $H$ and $W$, $\lambda = 1e-3, 1e-2, 1e-1$, respectively.}
\label{fig:cbcl}
\end{figure*}

\subsection{Learning a Basis for Face Images}
In this experiment, we use the proposed S-NMF and S-NMF-W frameworks to learn a basis for the CBCL face image dataset \footnote{Available at \url{http://cbcl.mit.edu/cbcl/software-datasets/FaceData.html}} \cite{hoyer2004non,lin2007convergence}. Each dataset image is a $19 \times 19$ grayscale image. We used $n = 3d$ and learned $W$ by running S-NMF with reweighted-$\ell_1$ regularization on $H$ and S-NMF-W with reweighted-$\ell_1$ regularization on $W$ and $H$. We used $\tau_w,\tau_h = 0.1$ and ran all algorithms to convergence. Due to a scaling indeterminacy, $W$ is normalized to have unit $\ell_2$ column norm at each iteration. A random subset of the learned basis vectors for each method with various levels of regularization is shown in Fig. \ref{fig:cbcl}. The results show the flexibility offered by the proposed framework. Fig. \ref{subfig:1}-\ref{subfig:3} show that decreasing $\lambda$ encourages S-NMF to learn high level features, whereas high values of $\lambda$ force basis vectors to resemble images from the dataset. Fig. \ref{subfig:4}-\ref{subfig:6} show a similar trend for S-NMF-W, although introducing a sparsity promoting prior on $\vect{W}$ tends to discourage basis vectors from resembling dataset images. It is difficult to verify Theorem \ref{thm:global convergence nmf} experimentally because $W$ must be normalized at each iteration to prevent scaling instabilities and there is no guarantee that a given stationary point $W^*$ has unit column norms. Nevertheless, the normalized KKT residual for the tested S-NMF-W algorithms with $W$ normalization at each iteration on the CSBL face dataset is reported in Table \ref{table:nmf kkt}.


\subsection{Computational Issues}
One of the advantages of using the proposed MUR's is that inference can be performed on the entire matrix simultaneously in each block of the block-coordinate descent procedure with relatively simple matrix operations. In fact, the computational complexity of the MUR's in \eqref{eq:mult rule rl2}, \eqref{eq:mult rule rl1}, \eqref{eq:mult rule block rl2}, and \eqref{eq:mult rule block rl1} is equivalent to that of the original NMF MUR given in \eqref{eq:MUR H} (which is $\mathcal{O}(nmr)$ where $r \leq \min (m,n)$ \cite{lin2007convergence}). In other words, the proposed framework allows for performing S-NNLS and S-NMF without introducing computational complexity issues. Another benefit of this framework is that the operations required are simple matrix-based computations which lend themselves to a graphics processing unit (GPU) implementation. For example, a 9-fold speed-up is achieved in computing 500 iterations of \eqref{eq:mult rule nmf} on a GPU compared to a CPU.

\section{Conclusion}
We presented a unified framework for S-NNLS and S-NMF algorithms. We introduced the RPESM as a sparsity promoting prior for non-negative data and provided details for a general class of S-NNLS algorithms arising from this prior. We showed that low-complexity MUR's can be used to carry out the inference, which are validated by a monotonicity guarantee. In addition, it was shown that the class of algorithms presented is guaranteed to converge to a set of stationary points, and that the local minima of the objective function are \textit{sparse}. This framework was then extended to a block coordinate descent technique for S-NMF and S-NMF-W. It was shown that the proposed class of S-NMF-W algorithms is guaranteed to converge to a set of stationary points.
\appendix
\section{Proof of Theorem \ref{thm:nmf}}
\label{appendix:proof nmf}
Due to the assumption on the form of $\pr{\Hij}$, the functional dependence of $\left\langle \left(\gamma_{(i,j)}\right)^{-z}\right\rangle$, and hence $\vOmijt$, on $\vHijt$ has the form $\left(\tau + \myexp{\vHijt}{z}\right)^{-1}$ up to a scaling constant, which is well-defined for all $\tau > 0$ and $\vHijt \in [0,\infty)$. As a result, \eqref{eq:mult rule nmf} is well defined for all $(i,j)$ such that $\vHijs{s} > 0$.

To show that $Q(H,\vHt)$ is non-increasing under MUR \eqref{eq:mult rule nmf}, a proof which follows closely to \cite{hoyer2002non,lee2001algorithms} is presented. We omit the $-\log u(\Hij)$ term in $Q(H,\vHt)$ in our analysis because it has no contribution to $Q(H,\vHt)$ if $H \geq 0$ and the update rules are guaranteed to keep $\Hij$ non-negative.

First, note that $Q(H,\vHt)$ is separable in the columns of $H$, $\Hj$, so we focus on minimizing $Q(H,\vHt)$ for each $\Hj$ separately. For the purposes of this proof, let $h$ and $x$ represent columns of $H$ and $X$, respectively, and let $Q(h)$ denote the dependence of $Q(H,\vHt)$ on one of the columns of $H$, with the dependency on $\vHt$ being implicit. Then,
\begin{align*}
Q\left(h\right) = \Vert x-W h\Vert_2^2 + \lambda \sum_{i}  q_i \myexp{{h_i}}{z}
\end{align*}
where $q$ represents the non-negative weights in \eqref{eq:Type 1 Q function}. Let $G(h,h^s)$ be
\begin{align}\label{eq:G}
\begin{split}
G(h,h^s) = Q(h^s) + (h-h^s)^T \triangledown Q(h^s)+ \frac{(h-h^s)^T {K}(h^s)(h-h^s)}{2}
\end{split}
\end{align}
where ${K}(h^s) = \diag{\left(W^T W h^s + \lambda z q \odot \myexp{h^s}{z-1}\right)/{h^s}}$.
For reference,
\begin{align}
\triangledown Q(h^s) &= W^T W h^s-W^T x + \lambda z q \odot \myexp{h^s}{z-1}\label{eq:first derivative}\\
\triangledown^2 Q(h^s) &= W^T W + \lambda z(z-1)\diag{q \odot \myexp{h^s}{z-2}} \label{eq:second derivative}.
\end{align}
It will now be shown that $G(h,h^s)$ is an auxiliary function for $Q(h)$. Trivially, $G(h,h) = Q(h)$. To show that $G(h,h^s)$ is an upper-bound for $Q(h)$, we begin by using the fact that $Q(h)$ is a polynomial of order $2$ to rewrite $Q(h)$ as $Q(h) = Q(h^s) + (h-h^s)^T \triangledown Q(h^s) +0.5{(h-h^s)^T \triangledown^2 Q(h^s)(h-h^s)}$.
It then follows that $G(h,h^s)$ is an auxiliary function for $Q(h)$ if and only if the matrix $M = {K}(h^s) - \triangledown^2 Q(h^s)$ is positive semi-definite (PSD). The matrix $M$ can be decomposed as $M = M_1 + M_2$, where $
M_1 = \diag{\left(W^T W h^s\right)/{h^s}}-W^T W$ and $M_2 = \lambda z(2-z) \diag{q \odot \myexp{h^s}{z-2}}$.
The matrix $M_1$ was shown to be PSD in \cite{lee2001algorithms}. The matrix $M_2$ is a diagonal matrix with the $(i,i)$'th entry being $\lambda z(2-z)q_i \myexp{{h_i^s}}{z-2}$. Since $q_i \myexp{{h_i^s}}{z-2} \geq 0$ and $z \leq 2$, $M_2$ has non-negative entries on its diagonal and, consequently, is PSD. Since the sum of PSD matrices is PSD, it follows that $M$ is PSD and $G(h,h^s)$ is an auxiliary function for $Q(h)$. Since $G(h,h^s)$ as an auxiliary function for $Q(h)$, $Q(h)$ is non-increasing under the update rule \cite{lee2001algorithms}
\begin{align}
\label{eq:aux obj}
{h^{s+1}} = \argmin_{h} G\left(h,h^s\right).
\end{align}
The optimization problem in \eqref{eq:aux obj} can be solved in closed form, leading to the MUR shown in \eqref{eq:mult rule nmf}. The multiplicative nature of the update rule in \eqref{eq:mult rule nmf} guarantees that the sequence $\lbrace \vHs{s} \rbrace_{s=1}^\infty$ is non-negative.

\section{Proof of Theorem \ref{thm:local}}
\label{appendix:proof of local}
This proof is an extension of (Theorem 1 \cite{subsetSelection}) and (Theorem 8 \cite{kreutz1997general}). Since $L(H)$ is separable in the columns of $H$, consider the dependence of $L(H)$ on a single column of $H$, denoted by $L(h)$. The function $L(h)$ can be written as
\begin{align}\label{eq:likelihood sparse}
\Vert x-Wh \Vert_2^2-
\sum_{i=1}^{n} 2\sigma^2 \log \p{h_i}.
\end{align}
Let $h^*$ be a local minimum of $L(h)$. We observe that $h^*$ must be non-negative. Note that $-\log \p{h_i} \rightarrow\infty$ when $h_i < 0$ since $\p{h_i} = 0$ over the negative orthant. As such, if one of the elements of $h^*$ is negative, $h^*$ must be a global maximum of $L(h)$. Using the assumption on the form of $\pr{h_i}$, \eqref{eq:likelihood sparse} becomes
\begin{align}\label{eq:likelihood sparse 2}
\Vert x-Wh \Vert_2^2+
\sum_{i=1}^{n} 2\sigma^2\left(\alpha \log \left(\tau+\myexp{h_i}{z}\right) - \log u\left(h_i\right)\right)+c
\end{align}
where constants which do not depend on $h$ are denoted by $c$. By the preceding argument, $\log u\left(h_i^*\right) = 0$, so the $\log u\left(h_i^*\right)$ term makes no contribution to $L(h^*)$. The vector ${h^*}$ must be a local minimum of the constrained optimization problem
\begin{align}\label{eq:basic}
\min_{x = W h + v^*}\underbrace{\sum_{i=1}^n  \log \left(\tau+\myexp{h_i}{z}\right)}_{\phi(h)}
\end{align}
where ${v^*} = x-W {h^*}$ and $\phi(\cdot)$ is the diversity measure induced by the prior on $\vect{H}$. It can be shown that $\phi(\cdot)$ is concave under the conditions of Theorem \ref{thm:local}. Therefore, under the conditions of Theorem \ref{thm:local}, the optimization problem \eqref{eq:basic} satisfies the conditions of (Theorem 8 \cite{kreutz1997general}). It then follows that the local minima of \eqref{eq:basic} are basic feasible solutions, i.e they satisfy $x = W h+{v^*}$ and $\Vert {h} \Vert_0 \leq d$. Since $h^*$ is one of the local minima of \eqref{eq:basic}, $\Vert {h^*} \Vert_0 \leq d$. 

\section{Proof of Theorem \ref{thm:rpesm coercive}}
\label{appendix:proof of coercive theorem}
It is sufficient to show that $
\lim_{\Hij \rightarrow \infty} \p{\Hij} = 0$.
Consider the form of $\p{\Hij}$ when it is a member of the RPESM family:
\begin{align}\label{eq:rpesm coercive}
\p{\Hij} = \int_0^\infty \p{\Hij | \gij} \p{\gij} d\gij
\end{align}
where $\vect{\Hij | \gij} \sim p^{RPE}(\Hij|\gij;z)$. Note that 
\begin{align*}
\vert p^{RPE}(\Hij | \gij) \p{\gij} \vert \leq \vert p^{RPE}(0|\gij;z) \p{\gij} \vert.
\end{align*}
Coupled with the fact that $\p{\Hij | \gij}$ is continuous over the positive orthant, the dominated convergence theorem can be applied to switch the limit with the integral in \eqref{eq:rpesm coercive}:
\begin{align*}
&\lim_{\Hij \rightarrow \infty} \int_0^\infty \p{\Hij | \gij} \p{\gij} d\gij \\
&=  \int_0^\infty \lim_{\Hij \rightarrow \infty}  \p{\Hij | \gij} \p{\gij} d\gij = 0.
\end{align*}

\section{Proof of Corollary \ref{lemma:existence of limit point}}
\label{appendix:proof of existence of limit point lemma}
This proof follows closely to the first part of the proof of (Theorem 1, \cite{zhao2016unified}). Let $\mathscr{S}_0 = \lbrace H \in \mathbb{R}_+^{n\times m} | L(H) \leq L(\bar{H}^0) \rbrace$. Lemma \ref{lemma:cost coercive} established that $L(H)$ is coercive. In addition, $L(H)$ is a continuous function of $H$ over the positive orthant. Therefore, $\mathscr{S}_0$ is a compact set (Theorem 1.2, \cite{burke}). The sequence $\lbrace L(\bar{H}^t) \rbrace_{t=1}^\infty$ is non-increasing as a result of Theorem \ref{thm:nmf}, such that $\lbrace \bar{H}^t \rbrace_{t=1}^\infty \in \mathscr{S}_0$. Since $\mathscr{S}_0$ is compact, $\lbrace \bar{H}^t  \rbrace_{t=1}^\infty$ admits at least one limit point.
\vspace{-1em}
\section{Proof of Theorem \ref{thm:global convergence}}
\label{appendix:proof of stationary-fixed point}
From Lemma \ref{lemma:existence of limit point}, the sequence $\lbrace \bar{H}^t \rbrace_{t=1}^\infty$ admits at least one limit point. What remains is to show that every limit point is a stationary point of \eqref{eq:Type1}. The sufficient conditions for the limit points to be stationary are (Theorem 1, \cite{wu1983convergence})
\begin{enumerate}
\item $Q(H,\vHt)$ is continuous in both $H$ and $\vHt$,
\item At each iteration $t$, one of the following is true
\begin{align}\label{eq:descendQ}
&\Q{\vHjs{t+1},\vHjt{t}} < \Q{\vHjt{t},\vHjt{t}}\\
\label{eq:minQ}
&\bar{H}_{(:,j)}^{t+1} = \argmin_{\Hj \geq 0 \footnotemark} \Q{\Hj,\vHjt{t}}.
\end{align}
\end{enumerate}
The function $Q(H,\vHt)$ is continuous in $H$, trivially, and in $\vHt$ if the functional dependence of $\pr{\vHijt}$ on $\vHijt$ has the form \eqref{eq:power function}.

\footnotetext{As in the proof of Theorem \eqref{thm:nmf}, we omit the $-\log u\left( \Hij \right)$ term from $\Q{\Hj,\vHjt{t}}$ and explicitly enforce the non-negativity constraint on $\Hj$.}

In order to show that the descent condition is satisfied, we begin by noting that $\Q{\Hj,\vHjt{t}}$ is strictly convex with respect to $\Hj$ if the conditions of Theorem \ref{thm:global convergence} are satisfied. This can be seen by examining the expression for the Hessian of $Q\left(\Hj,\vHjt{t}\right)$ in \eqref{eq:second derivative}. If $W$ is full rank, then $W^T W$ is positive definite. In addition, $\lambda z(z-1)\diag{\vOmjt \odot \myexp{\vHjs{s}}{z-2}}$ is PSD because $z \geq 1$. Therefore, the Hessian of $\Q{\Hj,\vHjt{t}}$ is positive definite if the conditions of Theorem \ref{thm:global convergence} are satisfied.

Since $S = 1$, ${\bar{H}_{(:,j)}^{t+1}}$ is generated by \eqref{eq:mult rule nmf} with $\vHs{s}$ replaced by $\vHt$. This update has two possibilities: (1) $\vHjt{t+1} \neq \vHjt{t}$ or
(2) $\vHjt{t+1} = \vHjt{t}$.
If condition (1) is true, then \eqref{eq:descendQ} is satisfied because of the strict convexity of $\Q{\Hj,\vHjt{t}}$ and the monotonicity guarantee of Theorem \ref{thm:nmf}.

It will now be shown that if condition (2) is true, then $\vHjt{t+1}$ must satisfy \eqref{eq:minQ}. Since $\Q{\Hj,\vHjt{t}}$ is convex, any $\vHjt{t+1}$ which satisfies the Karush-Kuhn-Tucker (KKT) conditions associated with \eqref{eq:minQ} must be a solution to \eqref{eq:minQ} \cite{boyd2004convex}. The KKT conditions associated with \eqref{eq:minQ} are given by \cite{lin2007convergence}:
\begin{align}
\Hij &\geq 0 \label{eq:KKT1}\\
\left(\triangledown \Q{\Hj,\vHjt{t}}\right)_i &\geq 0 \label{eq:KKT2}\\
\Hij \left( \triangledown \Q{\Hj,\vHjt{t}}\right)_i &= 0\label{eq:KKT3}
\end{align}
for all $i$. The expression for $\triangledown \Q{\Hj,\vHjt{t}}$ is given in \eqref{eq:first derivative}. For any $i$ such that $\vHijts{t+1} > 0$, $\iji{W^T X} = \iji{W^T W \bar{H}^{t+1}} + \lambda \vOmijt \myexp{\bar{H}_{(i,j)}^{t+1}}{z-1}$ because $\bar{H}^{t+1}$ was generated by \eqref{eq:mult rule nmf}. This implies that 
\begin{align*}
\left(\triangledown \Q{\Hj,\vHjt{t}} \bigg |_{\Hj = \vHjt{t+1}}\right)_i = 0
\end{align*}
for all $i$ such that $\vHijts{t+1} > 0$. Therefore, all of the KKT conditions are satisfied.

For any $i$ such that $\vHijts{t+1} = 0$, \eqref{eq:KKT1} and \eqref{eq:KKT3} are trivially satisfied. To see that \eqref{eq:KKT2} is satisfied, first consider the scenario where $z = 1$. In this case,
\begin{align*}
&\lim_{\vHijts{t+1} \rightarrow 0 } \left( \triangledown \Q{\Hj,\vHjt{t}} \bigg|_{\Hj = \vHjt{t+1}}\right)_i\\
&=^1 \lim_{\vHijts{t+1} \rightarrow 0 } \iji{W^T W\bar{H}^{t+1}} 
+ \frac{\lambda \myexp{\vHijts{t+1}}{0}}{\tau+\myexp{\vHijts{t+1}}{1}} - \iji{W^T X}\\
&= c+ \frac{\lambda}{\tau} - \iji{W^T X} \geq^2 0
\end{align*}
where $c \geq 0$, (1) follows from the assumption on $\pr{\Hij}$ having a power exponential form, and (2) follows from the assumptions that the elements of $W^T X$ are bounded and $\tau \leq \lambda / \max_{i,j}\iji{W^T X}$. When $z=2$, 
\begin{align*}
&\lim_{\vHijts{t+1} \rightarrow 0 } \lim_{\tau\rightarrow 0} \left( \triangledown \Q{\Hj,\vHjt{t}} \bigg|_{\Hj = \vHjt{t+1}}\right)_i\\
&=^1 \lim_{\vHijts{t+1} \rightarrow 0 } \iji{W^T W\bar{H}^{t+1}} 
+  \frac{2\lambda}{\vHijts{t+1}} - \iji{W^T X} \geq^2 0
\end{align*}
where (1) follows from the assumption on $\pr{\Hij}$ having a power exponential form and (2) follows from the assumption that the elements of $W^T X$ are bounded. Therefore, \eqref{eq:KKT2} is satisfied for all $i$ such that $\bar{H}_{(i,j)}^{t+1} = 0$. To conclude, if $\vHjt{t+1}$ satisfies $\vHjt{t+1} = \vHjt{t}$, then it satisfies the KKT conditions and must be solution of \eqref{eq:minQ}.
\vspace{-1em}
\section{Proof of Corollary \ref{cor:1}}
\label{appendix:proof of cor 1}
In the S-NMF setting ($\zeta_w = $ \eqref{eq:MUR H}, $\zeta_h = $ \eqref{eq:mult rule nmf}), this result follows from the application of (Theorem 1 \cite{lee2001algorithms}) to the $W$ update stage of Algorithm \ref{alg:Sparse NMF} and the application of Theorem \ref{thm:nmf} to the $H$ update stage of Algorithm \ref{alg:Sparse NMF}. In the S-NMF-W setting ($\zeta_w = $ \eqref{eq:mult rule nmf}, $\zeta_h = $ \eqref{eq:mult rule nmf}), the result follows from the application of Theorem \ref{thm:nmf} to each step of Algorithm \ref{alg:Sparse NMF}. In both cases,
\begin{align*}
L^{NMF}(\bar{W}^{t},\bar{H}^{t}) &\geq L^{NMF}(\bar{W}^{t+1},\bar{H}^{t}) \geq L^{NMF}(\bar{W}^{t+1},\bar{H}^{t+1}).
\end{align*}
\vspace{-3em}
\section{Proof of Corollary \ref{thm:global convergence nmf}}
\label{appendix:proof of thm 4}
The existence of a limit point $(\bar{W}^{\infty},\bar{H}^{\infty})$ is guaranteed by Corollary \ref{lemma:existence of limit point nmf}. It is sufficient to show that $L^{NMF}(\cdot,\cdot)$ is stationary with respect to $\bar{W}^{\infty}$ and $\bar{H}^\infty$ individually. The result follows by application of Theorem \ref{thm:global convergence} to $\bar{W}^\infty$ and $\bar{H}^\infty$.

\bibliography{IEEEabrv,../ManuscriptBib}

\end{document}